\documentclass[10pt,twocolumn,letterpaper]{article}

\usepackage[pagenumbers]{cvpr} 

\usepackage{multicol}
\usepackage{multirow}
\usepackage{tcolorbox}

\newcommand{\pldiff}[1]{\!\colorbox[cmyk]{0.1,0,0.1,0}{#1}\!}

\definecolor{cvprblue}{rgb}{0.21,0.49,0.74}
\usepackage[pagebackref,breaklinks,colorlinks,allcolors=cvprblue]{hyperref}

\def\paperID{xxxxx} 
\def\confName{CVPR}
\def\confYear{2026}

\title{SciPostGen: Bridging the Gap between Scientific Papers and Poster Layouts}
\author{
Shun Inadumi\textsuperscript{1,2}\thanks{Work done during an internship at OMRON SINIC X Corp.} \quad Shohei Tanaka\textsuperscript{1} \quad Tosho Hirasawa\textsuperscript{1}\\Atsushi Hashimoto\textsuperscript{1} \quad Koichiro Yoshino\textsuperscript{3,2} \quad Yoshitaka Ushiku\textsuperscript{1}\\
$^1$OMRON SINIC X Corp. \quad $^2$NAIST \quad $^3$The University of Osaka\\
{\tt\small inazumi.shun.in6@naist.ac.jp, shohei.tanaka@sinicx.com, tosho.hirasawa@sinicx.com} \\
{\tt\small atsushi.hashimoto@sinicx.com, yoshino.koichiro.es@osaka-u.ac.jp, yoshitaka.ushiku@sinicx.com} \\
}
\begin{document}
\maketitle
\begin{abstract}
As the number of scientific papers continues to grow, there is a demand for approaches that can effectively convey research findings, with posters serving as a key medium for presenting paper contents.
Poster layouts determine how effectively research is communicated and understood, highlighting their growing importance.
In particular, a gap remains in understanding how papers correspond to the layouts that present them, which calls for datasets with paired annotations at scale.
To bridge this gap, we introduce SciPostGen, a large-scale dataset for understanding and generating poster layouts from scientific papers.
Our analyses based on SciPostGen show that paper structures are associated with the number of layout elements in posters.
Based on this insight, we explore a framework, Retrieval-Augmented Poster Layout Generation, which retrieves layouts consistent with a given paper and uses them as guidance for layout generation.
We conducted experiments under two conditions: with and without layout constraints typically specified by poster creators.
The results show that the retriever estimates layouts aligned with paper structures, and our framework generates layouts that also satisfy given constraints.
The dataset and code are publicly available at \url{https://omron-sinicx.github.io/paper2layout/}.
\end{abstract}
    
\begin{figure}[t]
    \begin{minipage}[b]{\linewidth}
        \centering
        \includegraphics[keepaspectratio, width=\linewidth]{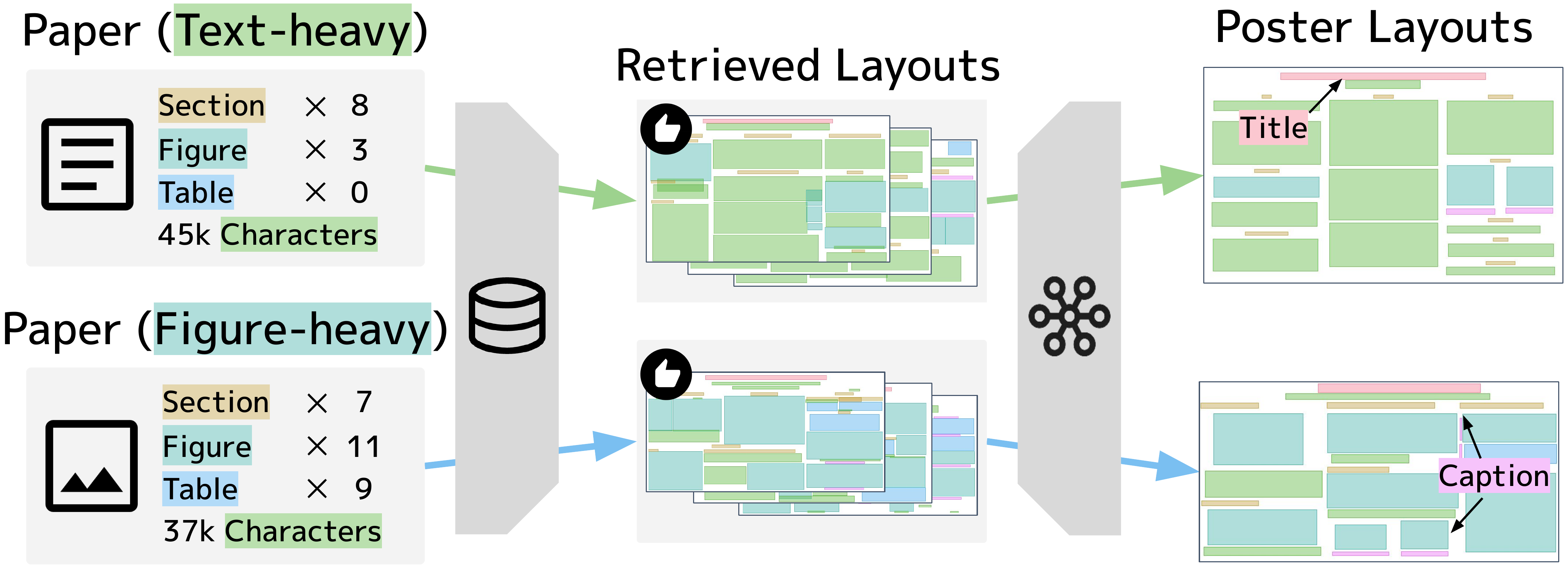}
        \subcaption{Concept of our framework}
        \label{fig:1_top_concept}
    \end{minipage}
    
    \vspace{1mm}
    
    \begin{minipage}[b]{\linewidth}
        \centering
        \includegraphics[keepaspectratio, width=\linewidth]{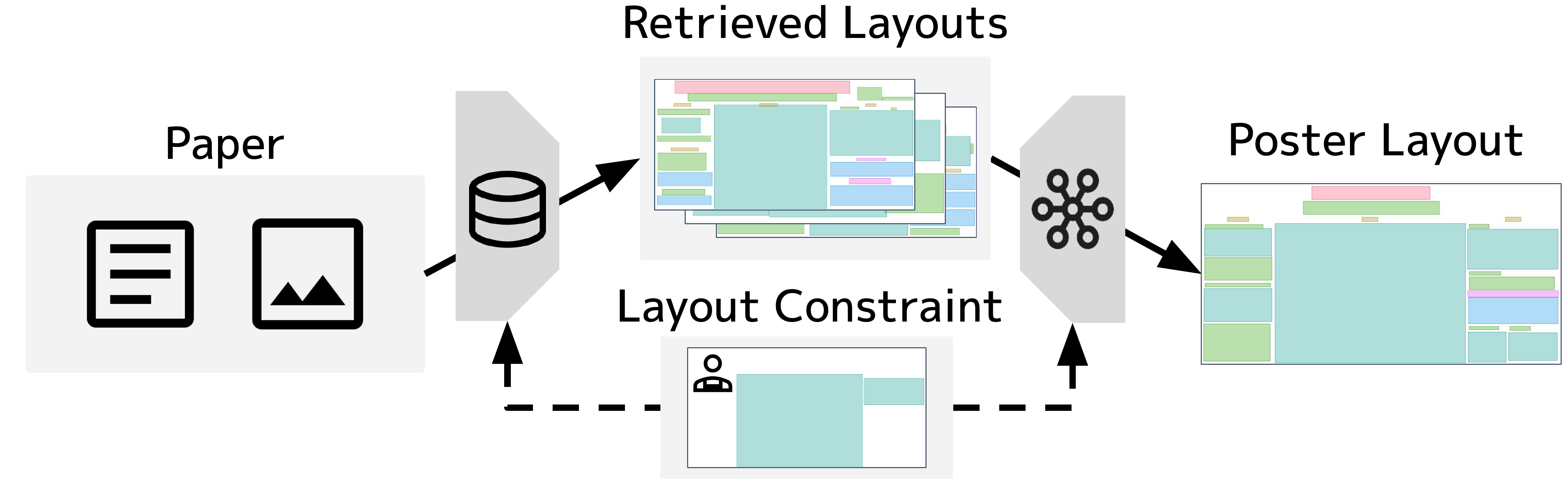}
        \subcaption{Experimental conditions: dashed arrows indicate optional data flows.}
        \label{fig:1_top_task-setting}
    \end{minipage}
    \vspace{-5mm}
    \caption{Overview of Retrieval-Augmented Poster Layout Generation:
    (a) Concept of our framework, where the layout retriever searches the SciPostGen dataset for poster layouts aligned with a given paper and provides them to the layout generator.
    (b) Experimental conditions, automatic and semi-automatic generation. In the semi-automatic setting, poster creators can optionally specify partial layouts as layout constraints.}
    \label{fig:1_top}
\end{figure}

\section{Introduction}
\label{sec:1_introduction}

The volume of research output has been increasing globally across disciplines, with the number of scientific papers continuing to grow over the long term~\cite{bornmann-and-mutz_2015, Bornmann-etal_2021}.
For example, the number of monthly submissions to arXiv, an open-access repository for scientific papers, increased from 7,983 in August 2015 to 21,825 in August 2025\footnote{\url{https://arxiv.org/stats/monthly_submissions }}.
As an enormous volume of research findings continues to be published each day, it has become difficult to keep track of individual papers, highlighting the growing need for systems that can efficiently summarize research output~\cite{Cohan-and-Goharian_2018, Yamamoto-etal_2021, Belouadi-etal-2024_ICLR, Gonzalez-etal_2024, Bandyopadhyay-etal_2024_INLG, Mondal-etal_2024-EACL, Ge-etal_2025_CVPR}.
One representative medium is the scientific poster, which visually summarizes key research findings and facilitates the communication of its main points within a limited time~\cite{Kanya-and-Leylia_2016, Qiang-etal_2016_AAAI, Qiang-etal_2019, Xu-Wan_2022_AAAI_demo}.

Generating posters from scientific papers has the potential to help researchers keep up with the growing volume of research outputs.
Achieving this requires addressing two complementary challenges: (i) a multimodal summarization problem, which concerns deciding what paper content should be included~\cite{Kanya-and-Leylia_2016, Jaisankar-etal_2025_AAAI, saxena-etal_2025_arXiv}, and (ii) a layout generation problem, which concerns deciding how paper content should be arranged~\cite{Qiang-etal_2016_AAAI, Qiang-etal_2019}.
Recent work has mainly focused on (i) and determines poster layouts using templates~\cite{Jaisankar-etal_2025_AAAI} or rules derived from paper structures~\cite{Pang-etal_2025_NeurIPS_benchmark, Sun-etal_2025_arXiv, zhang-etal_2025_arXiv}.
As layouts play an important role in how effectively the contents of papers are communicated and understood~\cite{Nelson-etal_1976, Larkin-and-Simon_1987}, the value of generating and evaluating layouts that align with papers has been recognized~\cite{Tanaka-etal_2024_BMVC, zhong-etal_2025}.
In particular, a gap remains in understanding how papers correspond to the layouts that present them, which calls for datasets with paired annotations at scale.

To bridge this gap, we introduce SciPostGen, a large-scale dataset that enables comprehensive analyses and learning of relationships between scientific papers and poster layouts.
Existing poster generation datasets contain only a few hundred paper–poster pairs, which are insufficient for studying such relationships in a data-driven manner~\cite{Qiang-etal_2016_AAAI, Qiang-etal_2019, Xu-Wan_2022_AAAI_demo, Tanaka-etal_2024_BMVC, Sun-etal_2025_arXiv}.
In contrast, SciPostGen has 18,097 paper-poster pairs.
Each paper is annotated with OCR text and bounding boxes of figures and tables, while each poster includes layout annotations for elements such as text blocks, figures, and captions.
We created these annotations through a combination of automatic processing and manual layout correction, ensuring both scalability and quality.

Our analyses based on SciPostGen show that paper structures are associated with the number of layout elements in their corresponding posters.
In particular, the amount of text and the number of figures and tables in a paper are moderately related to the number of text and figure elements in the layout.
These layout elements show slight correlations with the numbers and area coverage of their associated caption and section elements.
These findings suggest the possibility of generating poster layouts from papers.



Building on this insight, we explore a framework, Retrieval-Augmented Poster Layout Generation, which retrieves multiple layouts consistent with a given paper to accommodate the diversity observed in poster layouts, as shown in Figure~\ref{fig:1_top_concept}.
The layout retriever searches for layouts aligned with paper structures by learning relationships between papers and layouts through contrastive learning~\cite{Ting-etal_2020_ICML, Radford-etal_2021_ICML}.
Following recent work on layout generation~\cite{lin-etal_2023_NeurIPS, Seol-etal_2024_ECCV, Hsu-and-Peng_2025_CVPR}, we employ a large language model (LLM)~\cite{openai_2025_GPT5} as the layout generator to flexibly integrate diverse information, including retrieved results and paper structures. 
Additionally, our framework supports user-specified partial layouts used as layout constraints, as in~\cite{Gupta-etal_2021_ICCV, Li-etal_2020_arXiv}.

We evaluated our framework under two experimental conditions: automatic and semi-automatic poster generation settings, in which poster creators can provide layout constraints, as shown in Figure~\ref{fig:1_top_task-setting}.
In the semi-automatic setting, we simulate a practical workflow in which a creator places the main layout elements, and the system completes the remaining elements.
To implement these constraints, we take the two largest elements from a gold layout and use them as layout constraints.
Our experiments show that the layout retriever alone can estimate layouts aligned with paper structures.
Furthermore, our qualitative analysis suggests that our framework can refine retrieved layouts and produce layouts that are both consistent with paper structures and faithful to layout constraints.

The main contributions of this paper are as follows:
\begin{enumerate}
    \item We introduce the SciPostGen dataset, which surpasses existing datasets in scale and provides detailed annotations for paper and poster layout.
    \item Through analyses using SciPostGen, we clarify that paper structures are associated with the number and types of poster layout elements.
    \item We explore the Retrieval-Augmented Poster Layout Generation framework as a use case of SciPostGen.
    \item Our technical contribution lies in introducing a layout retriever that reflects the paper-to-layout relationships suggested by our analyses.
\end{enumerate}

\section{Related Work}
\label{sec:2_related_work}

\paragraph{Scientific Poster Generation.}
Driven by recent advances in LLMs, research on automatic poster generation from scientific papers has focused on LLM-based frameworks for generating poster content~\cite{Jaisankar-etal_2025_AAAI, Pang-etal_2025_NeurIPS_benchmark, zhang-etal_2025_arXiv, Sun-etal_2025_arXiv}.
For instance, \cite{Pang-etal_2025_NeurIPS_benchmark} incrementally generates poster content, yet its layout generation still relies on pre-defined rules.
This tendency is also observed in other studies~\cite{Jaisankar-etal_2025_AAAI, zhang-etal_2025_arXiv, Sun-etal_2025_arXiv}.
Our study focuses on generating poster layouts from scientific papers and highlights aspects less explored in existing work, which primarily emphasizes the multimodal summarization task.

Several datasets linking scientific papers with their corresponding posters have been developed, which can be used for poster layout generation~\cite{Qiang-etal_2016_AAAI, Qiang-etal_2019, Xu-Wan_2022_AAAI_demo, Tanaka-etal_2024_BMVC, zhong-etal_2025, Sun-etal_2025_arXiv}.
Early studies annotated correspondences between section, figure, and table counts in papers and coarse panel arrangements in posters~\cite{Qiang-etal_2019, Xu-Wan_2022_AAAI_demo}.
More recently, \cite{Tanaka-etal_2024_BMVC} has provided fine-grained annotations of poster layouts and a scheme.
However, because of the high labor cost of such annotations, the number of paper–poster pairs with comprehensive paper and layout annotations remains limited to only a few hundred\footnote{Although \cite{saxena-etal_2025_arXiv} contains around 16k paper--poster pairs, only poster images and paper abstracts are available.}.
To address this limitation, we constructed SciPostGen, a dataset that combines automatic annotation with manual layout correction to achieve both high-quality layout evaluation and scalable data collection.

\paragraph{Layout Generation.} 
Previous studies have explored layout generation in design domains such as web user interfaces~\cite{Deka-etal_2017_UIST}, advertising posters~\cite{Hsu-etal_2023_CVPR}, and document layout~\cite{Zhong-etal_2019_ICDAR}.
The layout generation task is typically categorized based on whether user-specified layout constraints~\cite{Li-etal_2018_ICLR, Li-etal_2020_arXiv, Arroyo_2021_CVPR}, such as the number of layout elements~\cite{Li-etal_2018_ICLR} or partial layouts~\cite{Li-etal_2020_arXiv}.
Across both settings, typical layout generation methods -- based on GANs~\cite{Li-etal_2018_ICLR, Li-etal_2021_TVCG, Kikuchi-etal_2021_ACMMM}, Transformers~\cite{Gupta-etal_2021_ICCV, Jiang-etal_2023_CVPR}, or Diffusion models~\cite{Inoue-etal_2023_CVPR, Chen-etal_2024_ICLR, Iwai-etal_2024_ECCV} -- take explicitly defined input representations to generate layouts.
In contrast to these methods, recent LLM-based approaches have demonstrated the ability to flexibly interpret unstructured inputs~\cite{lin-etal_2023_NeurIPS, Seol-etal_2024_ECCV, Hsu-and-Peng_2025_CVPR}.
Motivated by these developments, we employ an LLM to integrate retrieved layouts and paper structures within our framework. 

Retrieval-Augmented layout generation approaches have also been explored, where existing layout designs are retrieved and used to guide or constrain layout generation~\cite{Horita-etal_2024_CVPR, Wu-etal_2025_arXiv}.
For instance, \cite{Horita-etal_2024_CVPR} retrieves advertising poster layouts by treating the background image as a key cue that determines layout structure.
Such approaches leverage the diversity of existing layouts by retrieving multiple designs and conditioning a subsequent layout generator on them.
In the context of scientific poster layout generation, we view a paper as an implicit factor that conditions its layout.
Building on these perspectives, we explore a retrieval-augmented framework.
SciPostGen enables this through large-scale paired annotations of papers and poster layouts.


\begin{figure}[t]
    \centering
    \includegraphics[keepaspectratio, width=\linewidth]{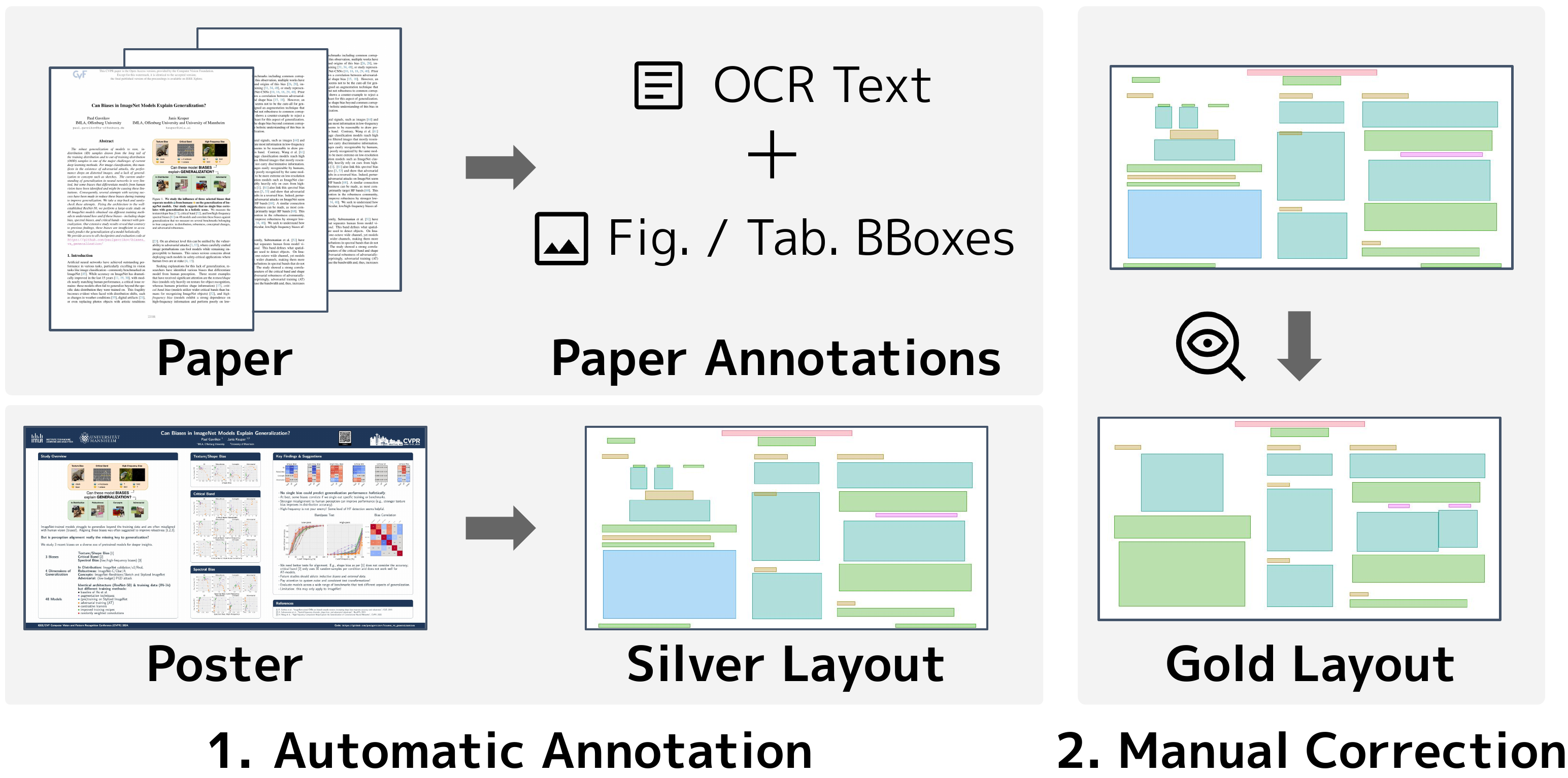}
    \caption{Annotation pipeline of SciPostGen: Paper and poster from \cite{Gavrikov-and-Keuper_2024_CVPR}, licensed under CC BY-SA 4.0.}
    \label{fig:3_annotation_pipeline}
\end{figure}


\section{SciPostGen Dataset}
\label{sec:3_dataset}
We constructed SciPostGen, a large-scale dataset consisting of 18,097 pairs of scientific papers and their corresponding posters.
SciPostGen includes pairs collected from major computer science conferences and covers both landscape and portrait poster formats.
The copyrights of collected papers and posters belong to the respective conferences and/or authors.
To avoid copyright issues, we release only download scripts for original paper PDFs and poster images, together with corresponding annotation files.

\subsection{Annotation Overview}
SciPostGen provides comprehensive annotations for both papers and posters.
Each paper is annotated with OCR text and bounding boxes (BBoxes) for figures and tables, while each poster is annotated with BBoxes for its layout elements.
We created these annotations through automatic processing for dataset scalability, followed by manual correction applied to a subset of layout annotations to ensure the quality of the evaluated layouts.

Following previous work~\cite{Tanaka-etal_2024_BMVC}, we defined eight categories of poster layout elements as follows:
\begin{tcolorbox}[colback=gray!10, colframe=white, boxrule=0pt, left=2mm, right=2mm, top=1mm, bottom=1mm]
\begin{itemize}
    \item \textbf{Title}: main title area of a poster.
    \item \textbf{Author Info}: areas of author names and affiliations.
    \item \textbf{Section}: section title areas.
    \item \textbf{Text}: paragraph areas merged into one BBox.
    \item \textbf{List}: list areas including bullet and numbered items.
    \item \textbf{Table}: table body areas.
    \item \textbf{Figure}: figure body areas.
    \item \textbf{Caption}: caption text areas for tables and figures.
\end{itemize}
\end{tcolorbox}
We excluded advertising information and institutional logos from the layout annotations, since they are not related to the content of papers.

\begin{figure}[t]
    \centering
    \includegraphics[width=\linewidth]{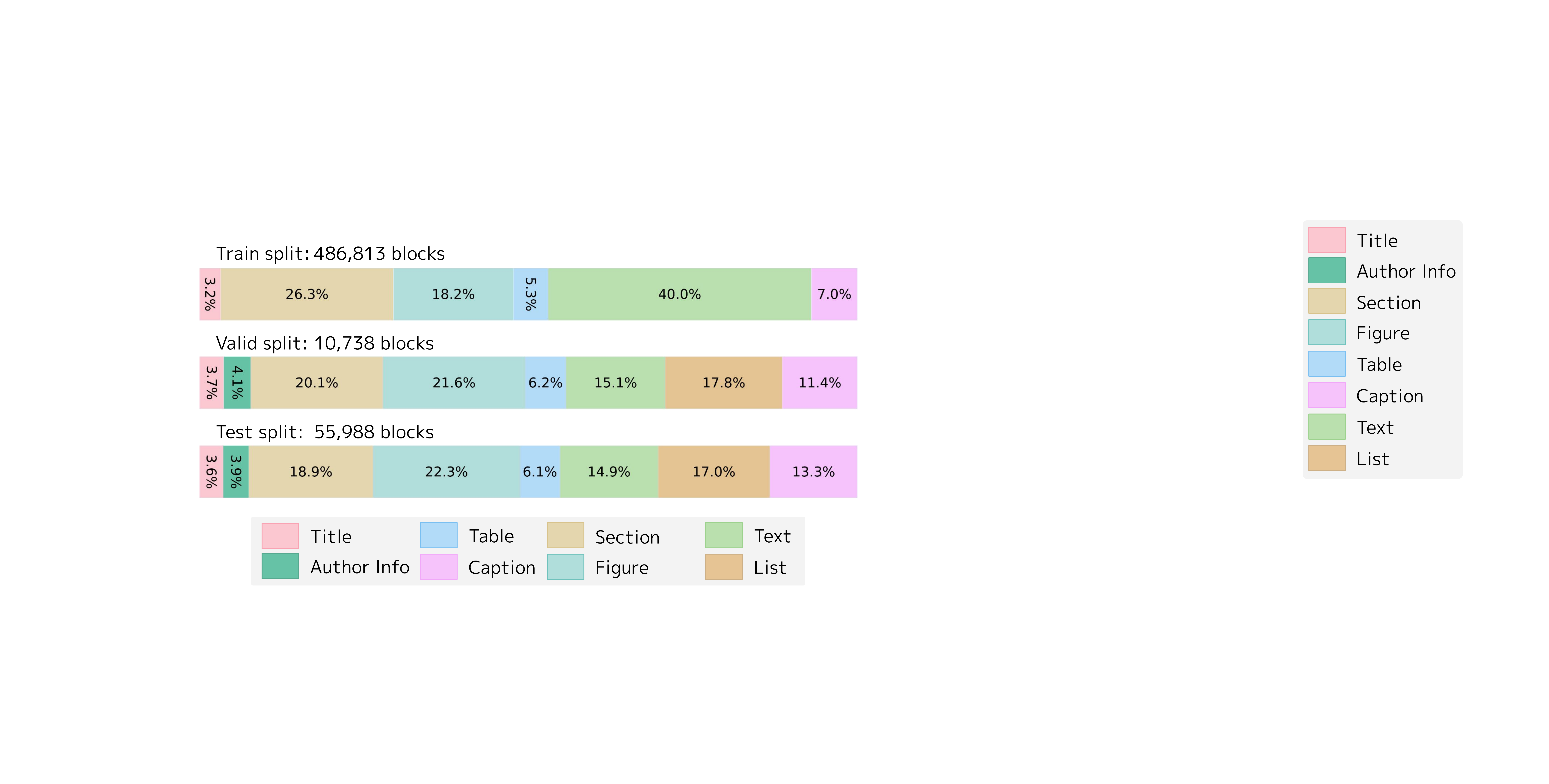}
    \caption{Layout element distributions}
    \label{fig:3_layout_dist}
\end{figure}

\begin{figure*}[t]
    \begin{minipage}[b]{0.51\linewidth}
        \centering
        \includegraphics[keepaspectratio, width=\linewidth]{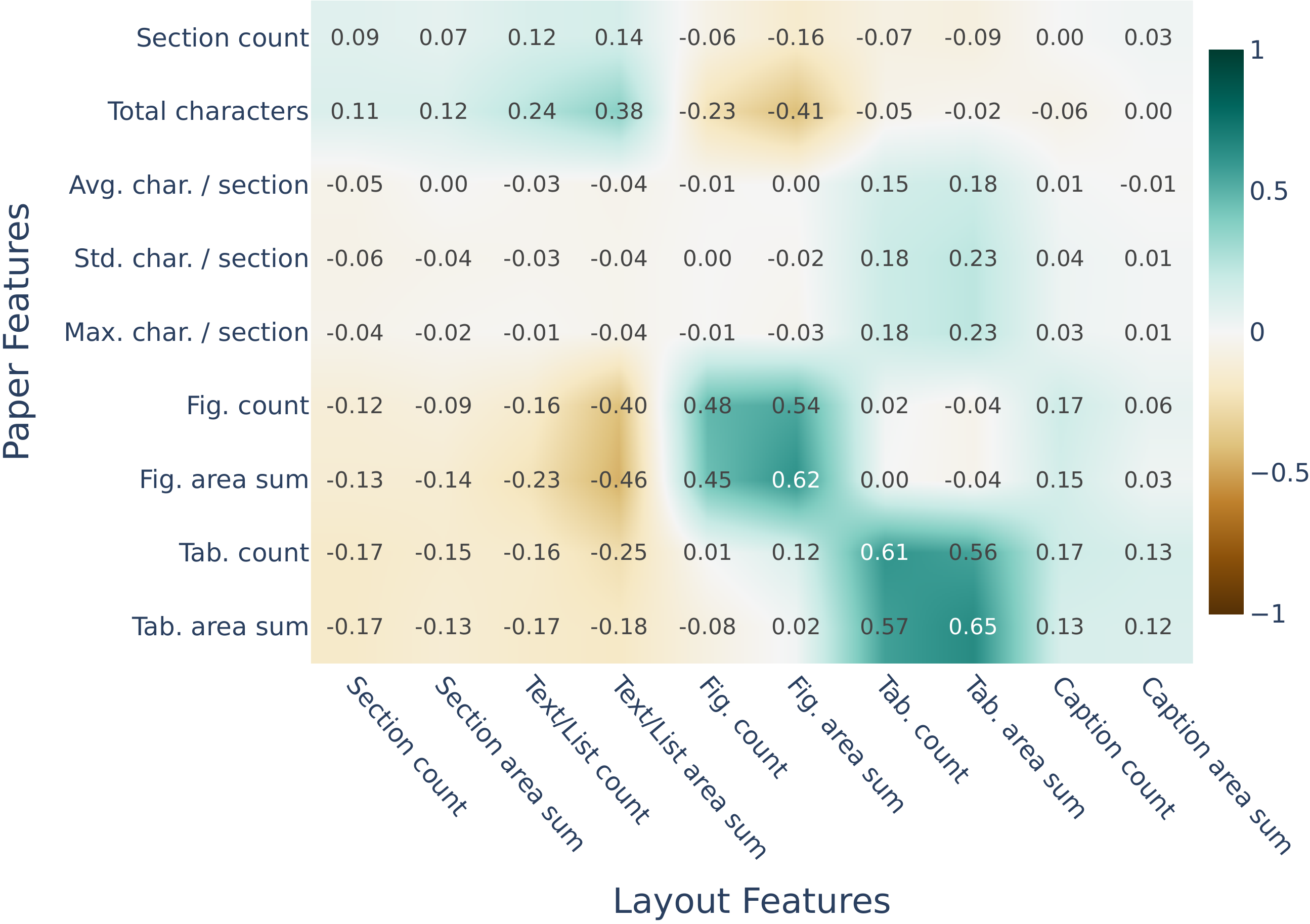}
        \subcaption{Spearman's $\rho$ between paper and layout features}
        \label{fig:3_paper-to-layout}
    \end{minipage}
    \begin{minipage}[b]{0.48\linewidth}
        \centering
        \includegraphics[keepaspectratio, width=\linewidth]{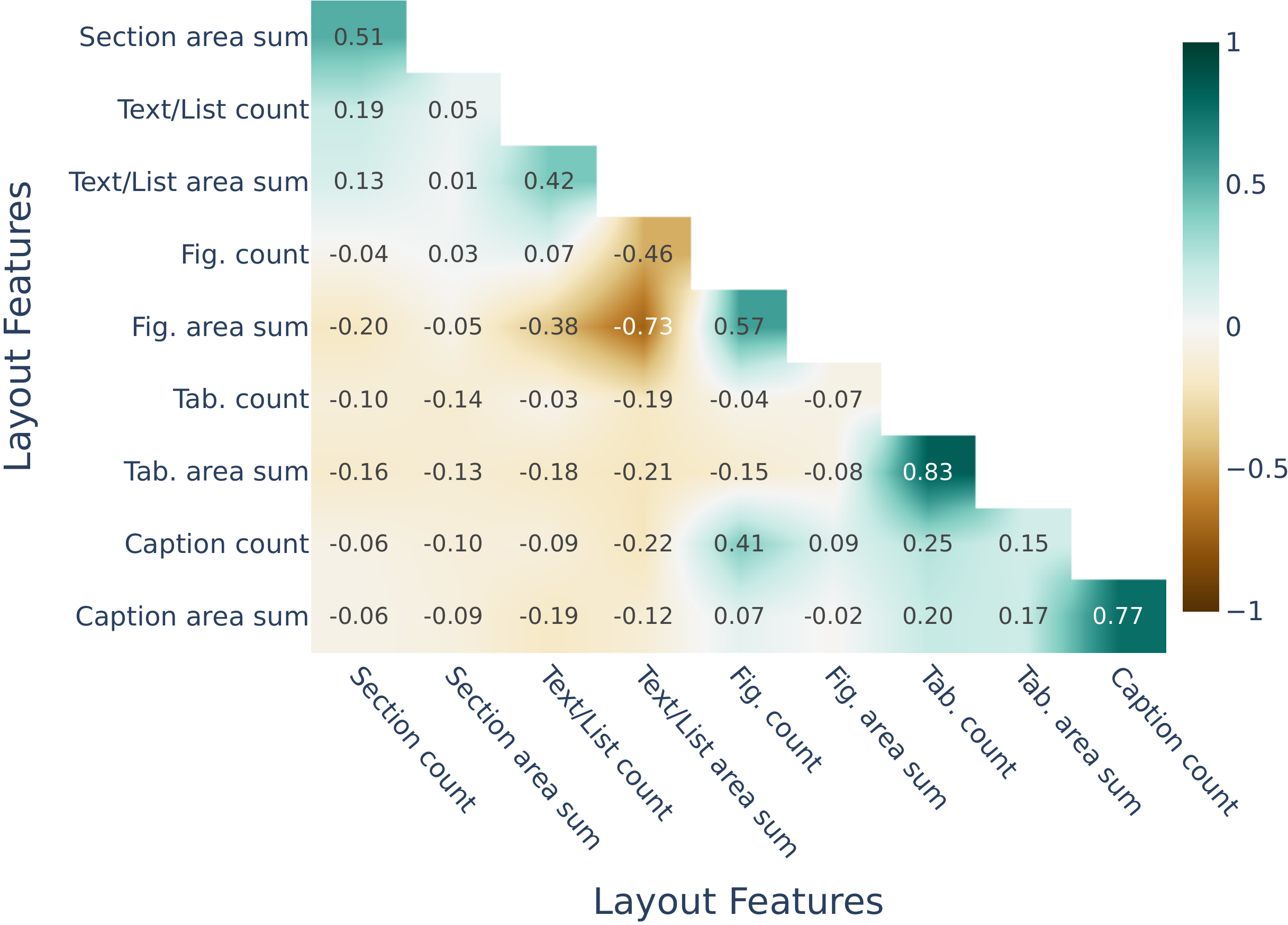}
        \subcaption{Spearman's $\rho$ between layout features}
        \label{fig:3_layout-to-layout}
    \end{minipage}
    \vspace{-2mm}
    \caption{Correlation analyses between paper structures and poster layouts in SciPostGen test split: Higher absolute Spearman's coefficients ($|\rho|$) indicate stronger correlations. ``List'' was merged into the ``Text,'' as they represent continuous textual content. ``Title'' and ``Author Info'' were excluded since each usually appears only once per poster.}
    \label{fig:3_corr-matrix}
\end{figure*}

\subsection{Data Construction}
We collected paper--poster pairs from the official websites of four major computer science conferences, CVPR, ICLR, ICML, and NeurIPS.
Figure~\ref{fig:3_annotation_pipeline} illustrates the annotation pipeline of SciPostGen, which consists of two stages: an automatic paper–poster annotation process and a manual layout correction process.

\paragraph{Automatic Paper--Poster Annotation.}
We generated paper and poster annotations using Nougat~\cite{Blecher-etal_2023_Nougat}, an OCR model for scientific documents, and Azure Document Intelligence\footnote{\url{https://azure.microsoft.com/products/ai-services/ai-document-intelligence}}, a commercial document analysis system.
For papers, we used an OCR model to convert PDF files into Markdown text and obtained figure and table BBoxes from Azure Document Intelligence.
For posters, we used Azure Document Intelligence to detect layout elements.
We then applied rule-based post-processing, such as merging adjacent text regions and removing noisy areas.
At this stage, we annotated five categories: Title, Section, Text, Table, and Caption, which we denote as the silver layouts\footnote{Azure Document Intelligence could not detect ``Author Info'' and ``List'' categories.}.

As a result of these processes, we obtained a dataset consisting of \textbf{15,710} training pairs and approximately 2,400 pairs for validation and testing.
To reduce bias across conferences, validation and test splits were randomly selected to maintain balance among conferences.

\paragraph{Manual Poster Layout Correction.}
We manually collected the silver layouts for the validation and test splits by professional annotators.
During this process, the silver layouts were refined to produce gold layouts, where layout elements that had been automatically detected as ``Text'' were further divided into ``Author Info,'' ``List,'' and ``Text.''
After excluding a few samples that were difficult to correct, we obtained \textbf{399} validation pairs and \textbf{1,988} test pairs.

\begin{figure*}[t]
    \centering
    \includegraphics[keepaspectratio, width=0.98\linewidth]{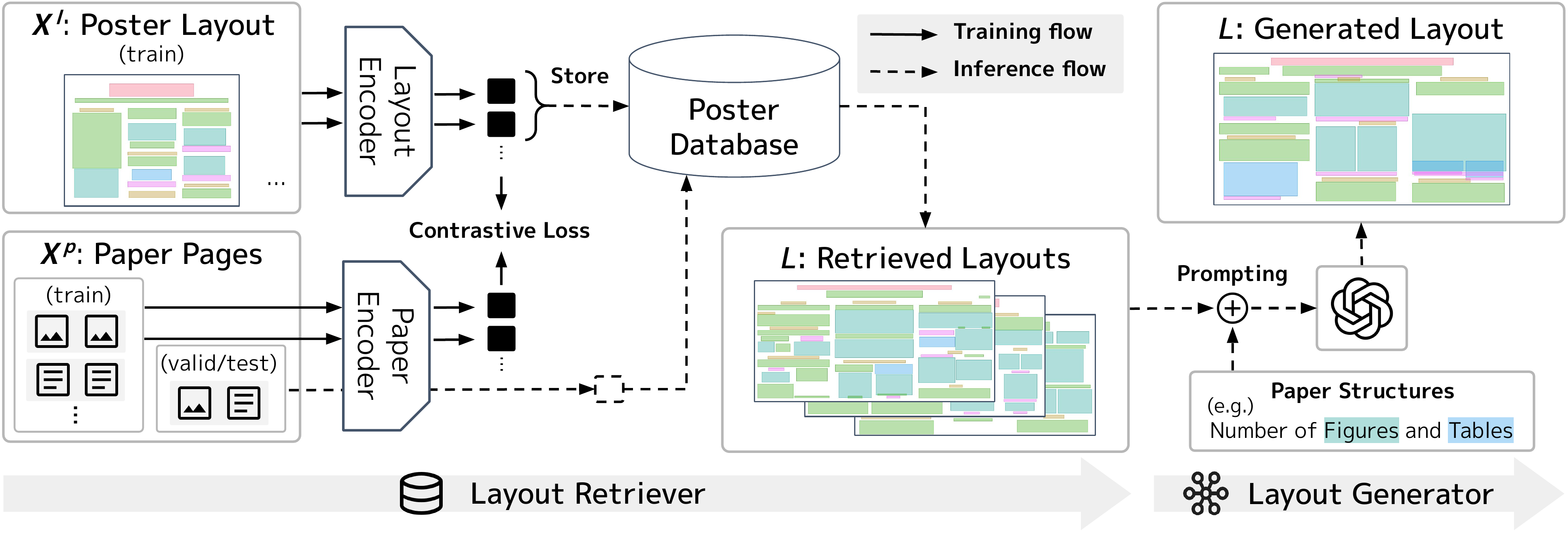}
    \vspace{-3mm}
    \caption{Overview of the Retrieval-Augmented Poster Layout Generation framework under the automatic setting.}
    \label{fig:4_framework_overview}
\end{figure*}

\subsection{Statistics and Analysis}
\label{sec:3_analysis}

\paragraph{Gold Layout vs. Silver Layout.}
We confirmed the differences between the silver and gold layouts using statistics of poster layout elements.
As shown in Figure~\ref{fig:3_layout_dist}, section elements occupy a larger proportion in the train split, while figures and tables are more frequent in the valid and test splits.
Captions are more common in the train split, whereas text elements appear more often in the valid and test splits.

\paragraph{Reliability of Silver Layout Annotations.}
To verify the consistency between the silver and gold layouts, we grouped layout elements into three categories: figures, tables, and other text elements.
We then computed the mean Average Precision across IoU thresholds between 0.50 and 0.95 at intervals of 0.05.
The resulting score was 0.53, which corresponds to a moderate level of agreement commonly observed in object detection benchmarks~\cite{Lin-etal_2014_ECCV, Redmon-etal_2016_CVPR}.
Based on this result, we consider that the layouts in the SciPostGen train split have sufficiently high annotation quality for training purposes.

\paragraph{Assessing Paper Structure and Layout Relationships.}
We quantitatively evaluated how paper structures relate to poster layouts using structural element-level features extracted from each pair.
On the paper side, we used features derived from OCR text length and the number and size of figures and tables.
On the layout side, we used those derived from the number and size of sections, figures, captions, and text--list elements.
We then computed Spearman rank correlation coefficients between the paired features using the SciPostGen test split, which provides manually corrected annotations and a sufficient number of samples for reliable correlation analysis.

Figure~\ref{fig:3_paper-to-layout} shows that paper structures, including text amount and the number of figures and tables, are moderately associated with the number of text, figure, and table elements in poster layouts.
For instance, papers with more text are commonly associated with layouts containing fewer figure elements, and vice versa, reflecting negative correlations between paper text amount or figure count and the corresponding layout figure or text area ($\rho < -0.40$).

\paragraph{Co-occurrence Patterns Among Layout Elements.}
As shown in Figure~\ref{fig:3_layout-to-layout}, caption elements show moderate positive correlations with figures ($\rho > 0.40$), while section counts exhibit weak positive correlations with textual elements ($\rho = 0.19$) and weak negative correlations with figure area ($\rho = -0.20$).
Taken together with the results in Figure~\ref{fig:3_paper-to-layout}, these patterns indicate that sections and captions show no clear associations with paper structures.
However, they exhibit slight to moderate correlations with major layout elements such as text and figures.
This suggests that paper characteristics may influence sections and captions indirectly through those major elements.



\section{Methodology}
\label{sec:4_methodology}
Figure~\ref{fig:4_framework_overview} shows our Retrieval-Augmented Poster Layout Generation framework, which consists of a layout retriever that searches for layouts aligned with a given paper and an LLM-based layout generator that produces a layout.
In this section, we describe the automatic poster generation setting, where a layout is generated directly from a given paper.
The semi-automatic setting with layout constraints is introduced in Section~\ref{sec:5_experimental_setup}.

\subsection{Task Definition}
We address the task of generating poster layouts from scientific papers.
Given a paper, the goal of this task is to output a poster layout $L = \{ (c_1, \mathbf{b}_1) \cdots \}$, where $c$ denotes a layout element category and $\mathbf{b} \in [0,1]^4$ represents a normalized BBox.
The input to the retriever is a sequence of $n_p$ page images converted from a paper PDF.
Each page is represented as an RGB image $\mathbf{X}^p \in \mathbb{R}^{n_p \times H \times W \times 3}$, where $H$ and $W$ denote the height and width of each page.
Paper structure can serve as auxiliary input to the layout generator, providing constraints such as the number of sections and the aspect ratios of figures and tables.

\subsection{Our Framework}
We first train the layout retriever with a contrastive learning objective~\cite{Ting-etal_2020_ICML, Radford-etal_2021_ICML}, where corresponding paper and layout pairs are treated as positives and other samples within the batch as negatives.
During inference, the retriever uses a paper in \textbf{test} split to retrieve the \textbf{top-3} layouts $L$ from \textbf{train} split using cosine similarity.
The layout generator then takes the retrieved layouts along with their corresponding paper structures as input to generate a layout.

\subsubsection{Layout Retriever}
The layout retriever consists of a paper encoder and a layout encoder, which are trained to retrieve layouts $L$ aligned with given paper pages $\mathbf{X}^p$.

\paragraph{Paper Encoder.}

Given a sequence of paper pages $\mathbf{X}^p$, the paper encoder outputs a $ d$-dimensional paper embedding $\mathbf{x}^p \in \mathbb{R}^d$ for layout retrieval, where $d$ denotes the embedding dimension.
Each page is first processed by a document image transformer, DiT~\cite{Li-etal_2022_ACMMM}, pretrained on document images, to extract patch-level features\footnote{We tested LayoutLMv3~\cite{Huang-etal_2022_ACMMM}, which jointly encodes document images and OCR tokens. However, we found it computationally inefficient for contrastive learning.}.
We apply attention pooling twice: the first stage aggregates patch-level features within each page, and the second stage aggregates the page-level features across the entire paper.
Finally, we pass the aggregated features through a two-layer MLP to obtain the paper embedding $\mathbf{x}^p$.

\paragraph{Layout Encoder.}
Given a layout RGB image $\mathbf{X}^l \in \mathbb{R}^{H \times W \times 3}$, the layout encoder outputs a layout embedding $\mathbf{x}^l \in \mathbb{R}^d$ for layout retrieval, where $H$ and $W$ are the same as for paper pages\footnote{Following existing work~\cite{Huang-etal_2022_ACMMM}, we experimented with using layout annotations $L$ directly as input. However, this approach showed lower retrieval performance compared to using images, such as $\mathbf{X}^l$.}.
The layout encoder is designed with an architecture similar to that of the paper encoder, consisting of a DiT backbone, attention pooling, and an MLP.
While both encoders follow a similar processing flow, the paper encoder uses two-stage attention pooling for multi-page aggregation, whereas the layout encoder applies one-stage pooling to a single layout image to obtain $\mathbf{x}^l$.

\paragraph{Loss Fuction.}

We train the layout retriever using a contrastive loss.
Given a batch of $N$ paper embeddings $\mathbf{x}^p_i$ and layout embeddings $\mathbf{x}^l_j$, the loss is defined as follows:
\begin{align}
    \mathcal{L} = -\frac{1}{N} \sum_{i=1}^N \log \frac{\exp(s_{ii})}{\sum_{j=1}^N \exp(s_{ij})},
\end{align}
where $s_{ij} = \frac{\cos(\mathbf{x}^p_i, \mathbf{x}^l_j)}{\tau}$ denotes the cosine similarity between $\mathbf{x}^p_i$ and $\mathbf{x}^l_j$, scaled by a temperature parameter $\tau$, and $\cos(\mathbf{x}^p_i, \mathbf{x}^l_j) = \frac{(\mathbf{x}^p_i)^\top \mathbf{x}^l_j}{|\mathbf{x}^p_i||\mathbf{x}^l_j|}$.

\subsubsection{Layout Generator}

The layout generator uses an LLM~\cite{openai_2025_GPT5} to generate a layout, taking as input the retrieved layouts and the corresponding paper structures.
In particular, these inputs are converted into HTML format~\cite{lin-etal_2023_NeurIPS, Seol-etal_2024_ECCV, Peng-etal_2024_ECCV} to construct prompts for the LLM, which then produces three layout candidates.
Following prior work~\cite{lin-etal_2023_NeurIPS}, we select the \textbf{top-1} layout based on reference-free metrics~\cite{Li-etal_2021_TVCG}, including overlap and alignment, as well as maximum IoU~\cite{Kikuchi-etal_2021_ACMMM} (Max. IoU).

\section{Experiments}
\label{sec:5_experiments}
\subsection{Experimental Setup}
\label{sec:5_experimental_setup}
\paragraph{Experimental Conditions.}
We consider two conditions: \textbf{Automatic poster generation}, where no layout constraints are given and a poster layout is generated directly from a paper; and \textbf{Semi-automatic poster generation}, where layout constraints are additionally provided as input to our framework.
The semi-automatic setting simulates a practical workflow in which a creator places the main layout elements and the system completes the remaining elements.
To define these constraints, we take the two largest elements from a gold layout and use them as layout constraints.

\paragraph{Compared Approaches.}
We compare our retriever to two baselines under the automatic and semi-automatic settings.
In the automatic setting, we randomly select three layouts from the retrieval pool without using the layout retriever, referred to as \textbf{Random 3-Samples}.
This baseline reflects the behavior of our framework when the retriever is disabled.
In the semi-automatic setting, we retrieve 15 layouts from the retrieval pool based on Max. IoU similarity to the partial layouts and use the top three layouts as a baseline, referred to as \textbf{Max. IoU Top-3}.
We compare this baseline with re-ranking by the retriever of Max. IoU top-15 layouts to examine whether the retriever improves layout quality.

We compare our framework, including the generator, to examine the role of paper structures as auxiliary inputs. 
In particular, we examine how these inputs influence the generated layouts compared to predictions obtained from the retriever alone. 
Note that a partial layout is additionally provided to the generator in the semi-automatic setting.

\paragraph{Evaluation Metrics.}
We evaluate predicted layouts by measuring BBox overlap and element count consistency, as poster layouts often contain many elements, for which count matching provides an insight into layout quality.
For BBoxes, we use mean IoU (mIoU), which is commonly used in layout generation tasks~\cite{Manandhar-etal_2020_ECCV, Bai-etal_2023_AAAI} to measure the agreement between predicted and gold layouts\footnote{Although Max. IoU is also widely used in layout generation tasks~\cite{lin-etal_2023_NeurIPS, Tanaka-etal_2024_BMVC}. We use mean IoU for fairness, since the generator uses Max. IoU.}.

For element counts, we use LTSim~\cite{Otani-etal_2024_arXiv}, which partially captures element type-level count consistency through IoU matching.
However, since LTSim does not provide an intuitive interpretation of element count consistency, we introduce two complementary metrics, $TC_{mean}$ and $TC_{std}$, where $TC$ stands for type-count.
These metrics quantify the mean and standard deviation of per-category differences in element counts,
which we formalize below.

Let $K$ be the set of element categories, and let $c_k^{gold}$ and $c_k^{pred}$ denote the number of elements in category $k \in K$ for a gold layout and a predicted layout, respectively.
We define $TC_{mean}$ and $TC_{std}$ as follows:
\begin{align}
    TC_{mean} &= \frac{1}{|K|} \sum_{k \in K} \Delta_k, \label{eq:tc_mean} \\
    TC_{std}  &= \sqrt{ \frac{1}{|K|} \sum_{k \in K} (\Delta_k - TC_{mean})^2 }, \label{eq:tc_std}
\end{align}
where $\Delta_k = c_k^{gold} - c_k^{pred}$ denotes the difference between the number of elements in the gold and predicted layouts for each category.
A \textbf{positive} $TC_{mean}$ indicates that the gold layout contains more elements than the predicted layout on average, whereas a \textbf{negative} value means that the predicted layout contains more.
$TC_{std}$ captures how uniformly these differences are distributed across element categories.

\subsection{Results under the Automatic Setting}
Table~\ref{tab:5_comparison_under_unconditional} shows results under the automatic setting, where (a) and (b) correspond to results obtained using GPT-5 and GPT-5-mini as the layout generator, respectively.

We focus on the case where paper structures are not provided as input (Approaches 2, 4, 6, and 7).
When the retrieved layouts were used as prompts for the generator, the generated results outperformed those obtained with Random 3-Samples in terms of mIoU, LTSim, and $TC_{std}$.
This finding suggests that, in LLM-based poster layout generation, higher-quality example layouts in prompts lead to better generated layouts.

We then compare results from the retriever alone with those obtained when its top-3 layouts are used as prompts for GPT-5-based generation (Approaches 3 and 4).
Combining the retriever with GPT-5 produces layouts that are closer to the gold layouts and show smaller variance across element categories, resulting in improvements in mIoU, LTSim, and $TC_{std}$, as shown in Table~\ref{tab:5_comp_uncond_gpt5}.
However, $TC_{mean}$ shows a degradation, indicating that generated layouts tend to include more elements than gold layouts.

When paper structures were additionally provided as auxiliary inputs to the generator (Approaches 4, 5, 7, and 8), GPT-5 achieved further improvements in mIoU compared with the setting without paper structures, as shown in Table~\ref{tab:5_comp_uncond_gpt5}.
In contrast, Table~\ref{tab:5_comp_uncond_gpt5-mini} shows that this improvement was not clearly observed when using GPT-5-mini.
These results suggest that a more capable LLM used as the generator can more effectively leverage paper structures.

\begin{table}[t]
    \begin{minipage}[t]{\linewidth}
        \centering
        \scalebox{0.70}{
            \begin{tabular}{ll|cccc}
                \toprule
                \# & \textbf{Approach} & \textbf{mIoU} $\uparrow$ & \textbf{LTSim} $\uparrow$ & \textbf{TC}$_{mean}$ $\rightarrow$ & \textbf{TC}$_{std}$ $\downarrow$ \\
                \cmidrule(lr){0-5}
                1 & Random 3 Samples (Avg.) & 0.122 & 0.637 & -0.548 & 3.643 \\
                2 & $\hookrightarrow$ w/o Paper Structures & 0.140  & 0.651 & 0.446 & 2.905 \\
                \cmidrule(lr){0-5}
                3 & Retrieved Top-3 (Avg.) & 0.145 & 0.651 & \textbf{-0.057} & 2.965 \\
                4 & $\hookrightarrow$ w/o Paper Structures & \pldiff{0.151} & \pldiff{\textbf{0.655}} & 0.619 & \pldiff{2.794} \\
                5 & $\hookrightarrow$ w/\phantom{o} Paper Structures & \pldiff{\textbf{0.159}} & 0.642 & 0.228 & 3.128 \\
                \bottomrule
            \end{tabular}
        }
        \vspace{0.5mm}
        \subcaption{\textbf{GPT-5}}
        \label{tab:5_comp_uncond_gpt5}
    \end{minipage}
    
    \vspace{1mm}

    \begin{minipage}[t]{\linewidth}
        \centering
        \scalebox{0.70}{
            \begin{tabular}{ll|cccc}
                \toprule
                \# &\textbf{Approach} &\textbf{mIoU} $\uparrow$ & \textbf{LTSim} $\uparrow$ & \textbf{TC}$_{mean}$ $\rightarrow$ & \textbf{TC}$_{std}$ $\downarrow$ \\
                \cmidrule(lr){0-5}
                1 & Random 3 Samples (Avg.) & 0.122 & 0.637 & -0.548 & 3.643 \\
                6 & $\hookrightarrow$ w/o Paper Structures & 0.138 & 0.640 & 0.384 & 2.958 \\
                \cmidrule(lr){0-5}
                3 & Retrieved Top-3 (Avg.) & 0.145 & \textbf{0.651} & \textbf{-0.057} & 2.965 \\
                7 & $\hookrightarrow$ w/o Paper Structures & 0.145 & 0.645 & 0.609 & \pldiff{\textbf{2.832}} \\
                8 & $\hookrightarrow$ w/\phantom{o} Paper Structures & \pldiff{\textbf{0.148}} & 0.630 & -0.759 & 3.638 \\
                \bottomrule
            \end{tabular}
        }
        \vspace{0.5mm}
        \subcaption{\textbf{GPT-5-mini}}
        \label{tab:5_comp_uncond_gpt5-mini}
    \end{minipage}
    \vspace{-3mm}
    \caption{Comparison of approaches under the \textbf{automatic generation} setting. \pldiff{Highlighted values} show improvements over retrieved results.}
    \label{tab:5_comparison_under_unconditional}
\end{table}

\begin{table}[t]    
    \begin{minipage}[t]{\linewidth}
        \centering
        \scalebox{0.70}{
            \begin{tabular}{ll|cccc}
                \toprule
                \# & \textbf{Approach} & \textbf{mIoU} $\uparrow$ & \textbf{LTSim} $\uparrow$ & \textbf{TC}$_{mean}$ $\rightarrow$ & \textbf{TC}$_{std}$ $\downarrow$ \\
                \cmidrule(lr){0-5}
                1 & Max. IoU Top-3 (Avg.) & 0.192 & 0.658 & 0.297 & 3.003 \\
                2 & $\hookrightarrow$ w/\phantom{o} Paper Structures & \pldiff{0.235} & \pldiff{0.668} & 0.717 & 2.931 \\
                \cmidrule(lr){0-5}
                3 & Retrieved Top-3 (Avg.) & 0.196 & 0.662 & \textbf{0.194} & 2.873 \\
                4 & $\hookrightarrow$ w/o Paper Structures & \pldiff{0.231} & \pldiff{0.670} & 1.316 & \pldiff{\textbf{2.824}} \\
                5 & $\hookrightarrow$ w/\phantom{o} Paper Structures & \pldiff{\textbf{0.238}} & \pldiff{0.668} & 0.664 & 2.941 \\
                \bottomrule
            \end{tabular}
        }
        \vspace{0.5mm}
        \label{tab:5_comp_cond_gpt5}
        \subcaption{\textbf{GPT-5}}
    \end{minipage}
    
    \vspace{1mm}

    \begin{minipage}[t]{\linewidth}
        \centering
        \scalebox{0.70}{
            \begin{tabular}{ll|cccc}
                \toprule
                \# & \textbf{Approach} & \textbf{mIoU} $\uparrow$ & \textbf{LTSim} $\uparrow$ & \textbf{TC}$_{mean}$ $\rightarrow$ & \textbf{TC}$_{std}$ $\downarrow$ \\
                \cmidrule(lr){0-5}
                1 & Max. IoU Top-3 (Avg.) & 0.192 & \textbf{0.658} & 0.297 & 3.003 \\
                6 & $\hookrightarrow$ w/\phantom{o} Paper Structures & \pldiff{0.211} & 0.642 & -0.712 & 3.558 \\
                \cmidrule(lr){0-5}
                3 & Retrieved Top-3 (Avg.) & 0.196 & 0.662 & \textbf{0.194} & 2.873 \\
                7 & $\hookrightarrow$ w/o Paper Structures & \pldiff{\textbf{0.225}} & 0.659 & 1.526 & \pldiff{\textbf{2.836}} \\
                8 & $\hookrightarrow$ w/\phantom{o} Paper Structures & \pldiff{0.215} & 0.643 & -0.659 & 3.527 \\
                \bottomrule
            \end{tabular}
        }
        \vspace{0.5mm}
        \label{tab:5_comp_cond_gpt-5-mini}
        \subcaption{\textbf{GPT-5-mini}}
    \end{minipage}
    \vspace{-3mm}
    \caption{Comparison of approaches under the \textbf{semi-automatic generation} setting, where allows layout constraints as inputs. \pldiff{Highlighted values} show improvements over retrieved results.}
    \label{tab:5_comparison_under_conditional}
\end{table}

\subsection{Results under the Semi-Automatic Setting}

Table~\ref{tab:5_comparison_under_conditional} shows results under the semi-automatic setting.
We compare Max. IoU Top-3 and the case where layouts are re-ranked by the retriever (Approaches 1 and 3).
When the Max. IoU top-15 layouts were re-ranked by the retriever all evaluation metrics improved.
In particular, the notable improvements in $TC_{mean}$ and $TC_{std}$ indicate that the retrieval module tends to select layouts that are more balanced across element categories and have a more appropriate number of elements similar to the gold layouts.

We compare the conditions where layout constraints are provided with and without additional paper structures (Approaches 4, 5, 7, and 8).
As shown in Table~\ref{tab:5_comparison_under_conditional}, a notable improvement in mIoU was observed only when GPT-5 was used as the generator.
This trend is consistent with the pattern observed under the unconditional setting.

\begin{figure}[t]
    \begin{minipage}[b]{\linewidth}
        \centering
        \includegraphics[keepaspectratio, width=\linewidth]{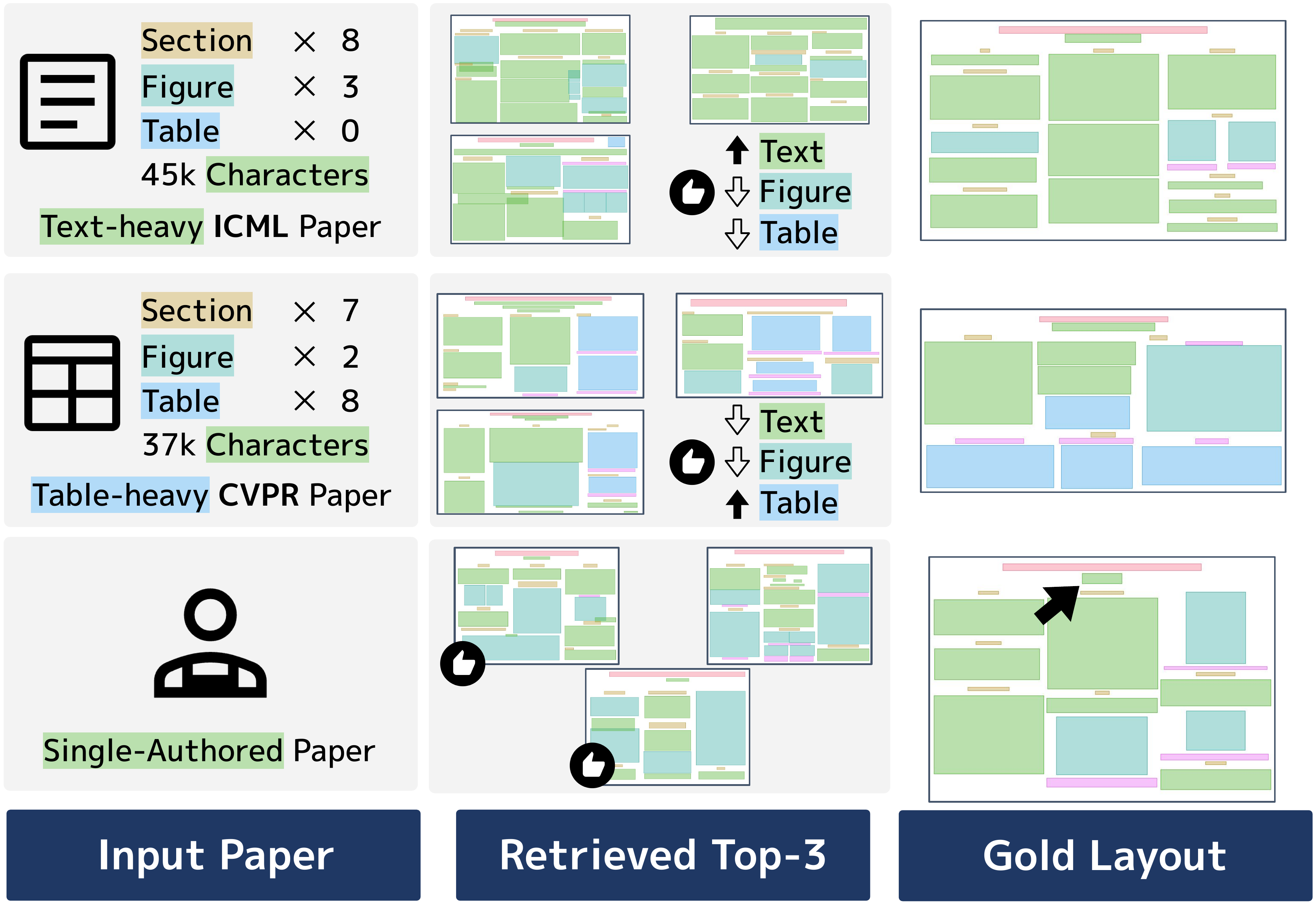}
        \subcaption{Top: from~\cite{Sharma-and-Amit_2024_ICML}; Middle: from~\cite{Park-and-Byun_2024_CVPR}; Bottom: from~\cite{Wouters_2024_ICML}}
        \label{fig:5_uncond_ret_examples}
    \end{minipage}

    \vspace{1mm}
    
    \begin{minipage}[b]{\linewidth}
        \centering
        \includegraphics[keepaspectratio, width=\linewidth]{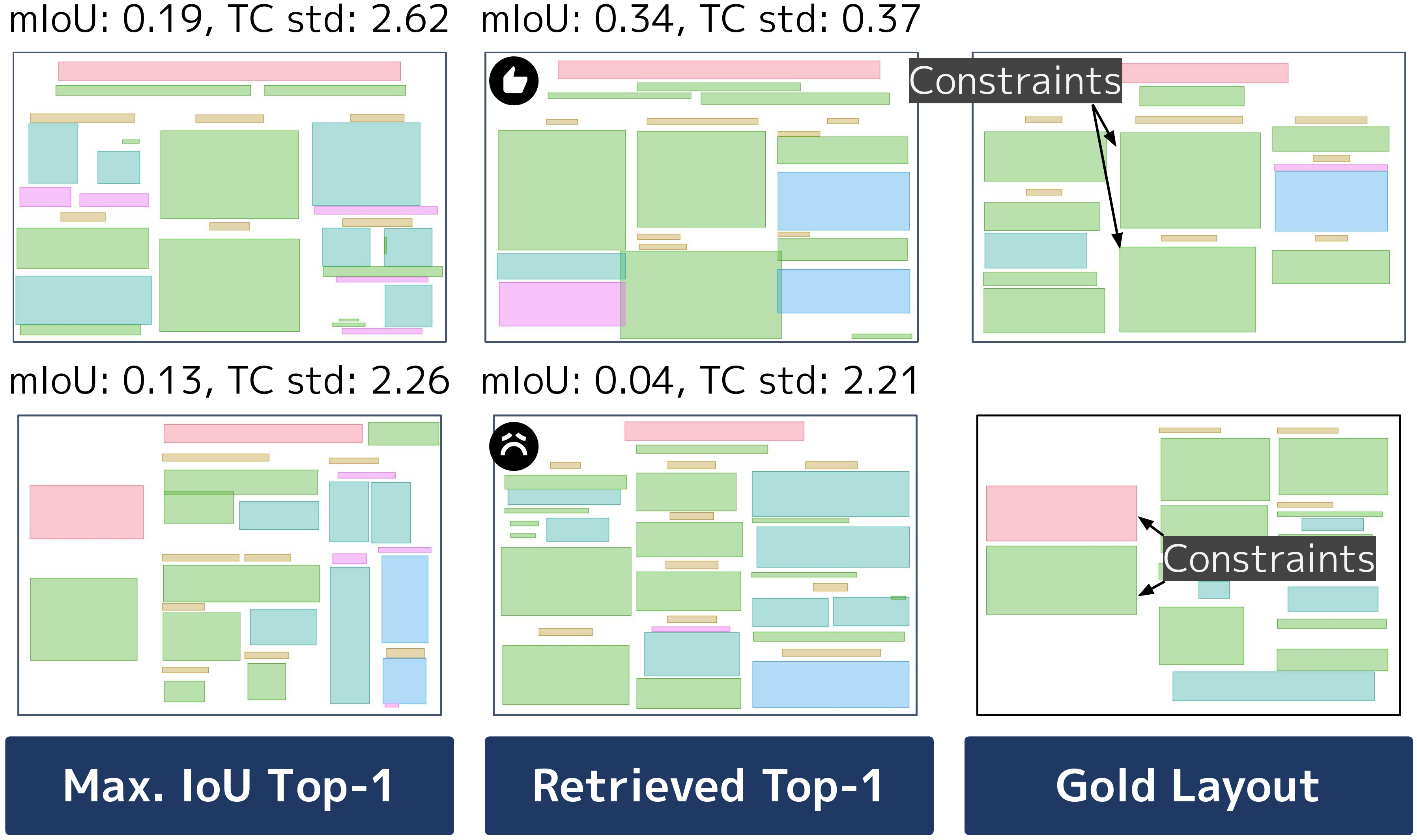}
        \subcaption{Top: from~\cite{Liu-etal_2024_ICML}; Bottom: from~\cite{Baker-etal_2024_NeurIPS}}
        \label{fig:5_cond_ret_examples}
    \end{minipage}
    \vspace{-5mm}
    \caption{Examples of retrieved results under the (a) \textbf{automatic} and (b) \textbf{semi-automatic} settings}
    \label{fig:5_retrieved_examples}
\end{figure}

\begin{figure}[t]
    \begin{minipage}[b]{\linewidth}
        \centering
        \includegraphics[keepaspectratio, width=\linewidth]{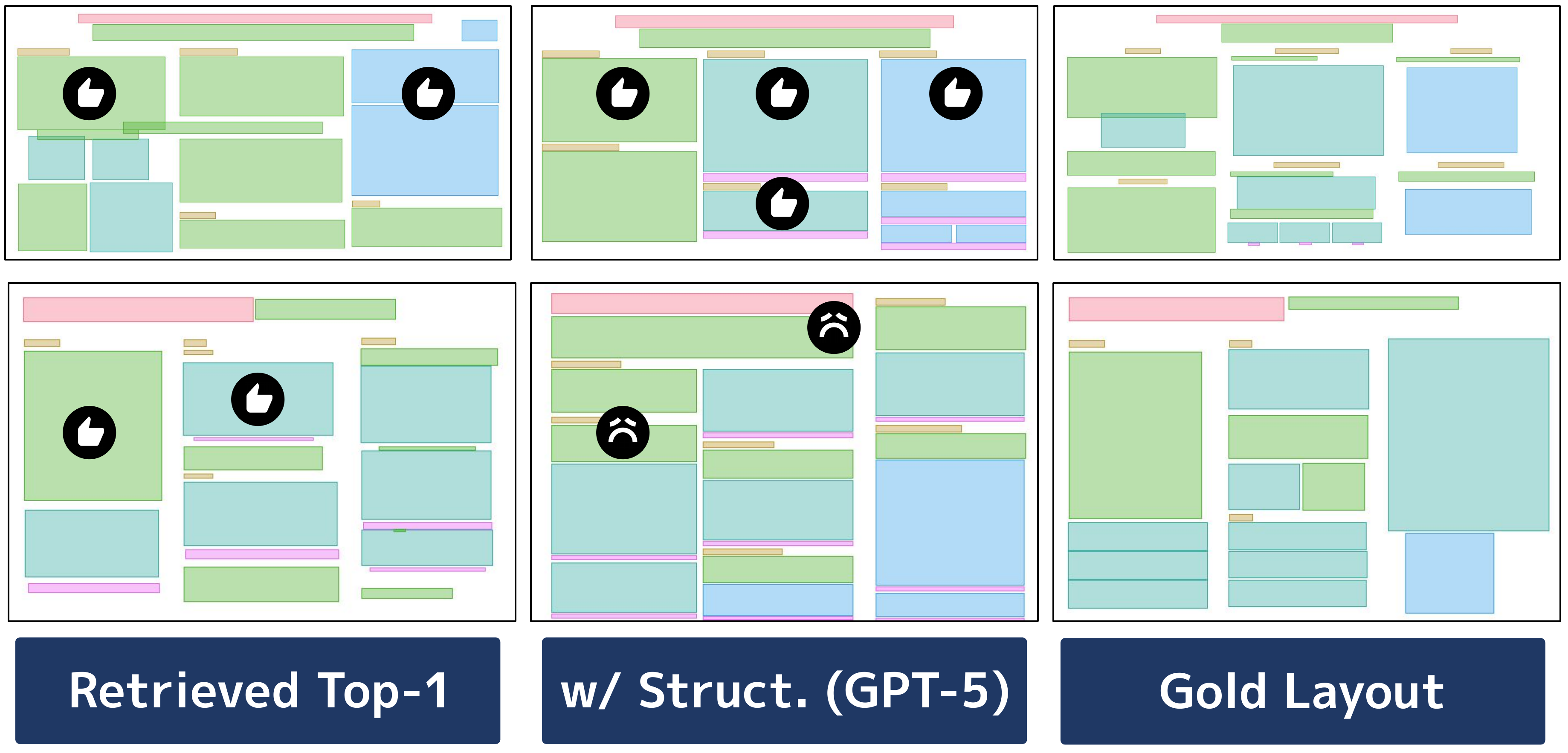}
        \subcaption{Top: from \cite{Yang-etal_2024_CVPR}; Bottom: from \cite{Zhang-etal_2024_NeurIPS}}
        \label{fig:5_uncond_framework_examples}
    \end{minipage}

    \vspace{1mm}
    
    \begin{minipage}[b]{\linewidth}
        \centering
        \includegraphics[keepaspectratio, width=\linewidth]{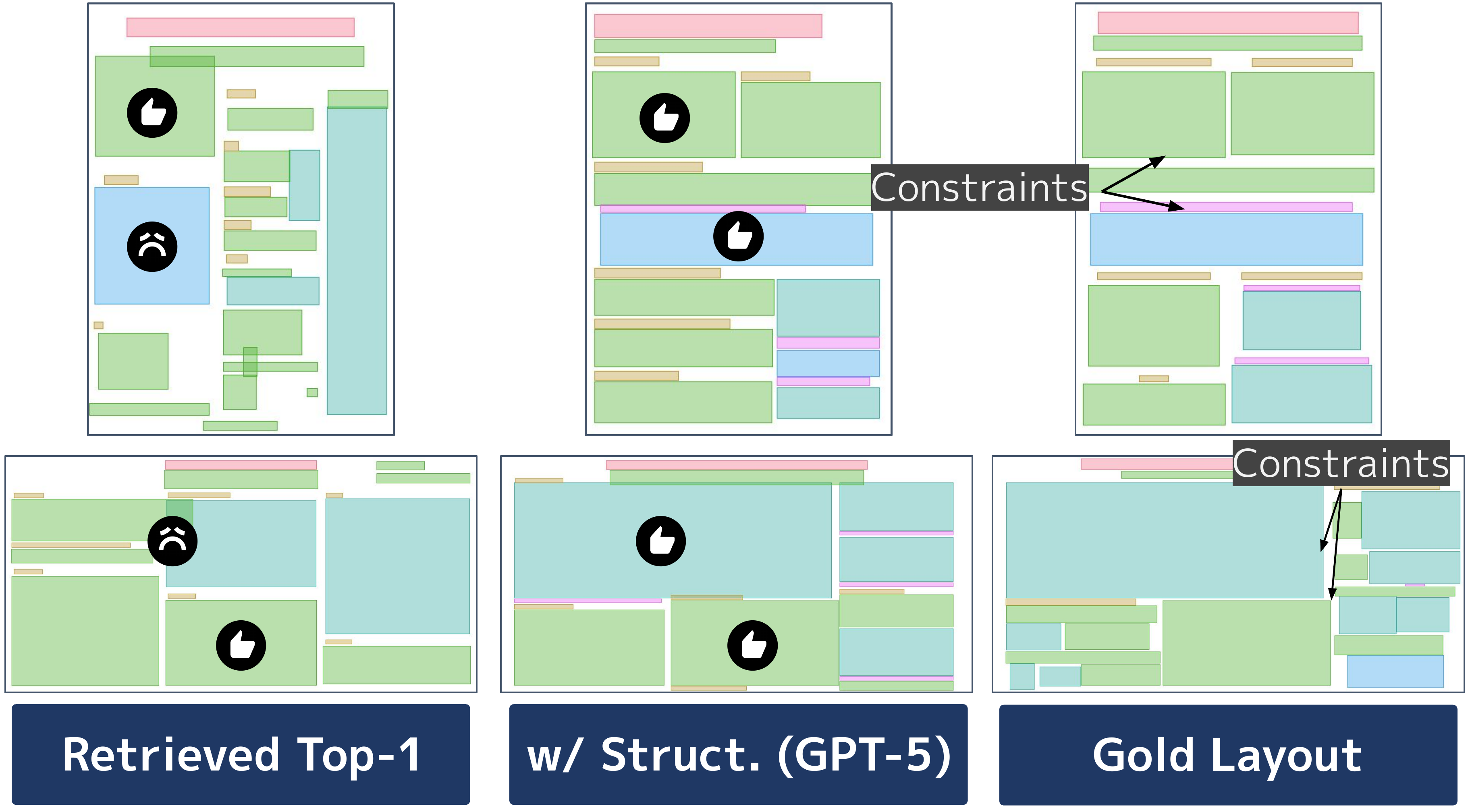}
        \subcaption{Top: from \cite{Mangold-etal_2024_NeurIPS}; Bottom: from \cite{Poduval-etal_2024_CVPR}}
        \label{fig:5_cond_framework_examples}
    \end{minipage}
    \vspace{-5mm}
    \caption{Examples of generation results under the (a) \textbf{automatic} and (b) \textbf{semi-automatic} settings}
    \label{fig:5_framework_examples}
\end{figure}

\subsection{Qualitative Analysis and Discussions}
\paragraph{Retrieved Results.}

Figure~\ref{fig:5_retrieved_examples} presents examples of the retrieved layouts.
As shown in Figure~\ref{fig:5_uncond_ret_examples}, the retrieved layouts are largely consistent with the gold layouts in terms of the number of text, figure, and table elements.
Notably, even for CVPR papers, which tend to include more figures than papers from other conferences, the retrieved layouts show a comparable number of layout elements to the gold ones.
This example suggests that the retriever is not significantly affected by biases inherent to specific paper formats.

When a single-authored paper was used as input, the gold layout showed a small proportion of the poster canvas occupied by the ``Author Info'' category.
In this case, two out of the top-3 retrieved layouts were consistent with the gold layout.
These findings from Figure~\ref{fig:5_uncond_ret_examples} indicate that the retriever alone can \textbf{appropriately retrieve layouts that reflect the structures of input papers}.

We compare examples between Max. IoU Top-3 and layouts re-ranked by the retriever.
The upper part of Figure~\ref{fig:5_cond_ret_examples} shows a successful example of re-ranking.
Owing to the characteristics of the retriever mentioned above, the retriever selected a layout consistent with the input paper and satisfying the given layout constraint, leading to improvements in both mIoU and $TC_{std}$.
Specifically, the retrieved layout corresponds to a gold layout that contains a few figures and a single table.
The lower part of Figure~\ref{fig:5_cond_ret_examples} shows an error case of re-ranking, where applying the retriever caused the layout constraints to be violated.
This type of layout, which allows greater flexibility in element arrangement, rarely appears in the retrieval pool, which corresponds to the SciPostGen train split, making it difficult for the retriever alone to select an appropriate layout.

\paragraph{Generation Results.}

Figure~\ref{fig:5_framework_examples} presents examples of poster layouts generated by our framework.
The upper part of Figure~\ref{fig:5_uncond_framework_examples} shows a case where the retrieved layouts contributed to the generator, resulting in improved quality of the generated layouts.
At the retrieval stage, the text block in the upper-left and the table on the right closely matched those in the gold layout.
When the paper constraints were additionally provided, multiple figures with aspect ratios consistent with the gold layout were added in the center.

The lower part of Figure~\ref{fig:5_uncond_framework_examples} shows an error case where adding the paper constraints degraded the generation quality.
At the retrieval stage, results included regions like the title and author information that matched those in the gold layout.
However, when the paper constraints were provided, these elements were excluded from the generated layout.

Figure~\ref{fig:5_cond_framework_examples} demonstrates examples where the generator successfully produced layouts that satisfied the given layout constraints.
Although the re-ranked layouts from the retriever did not fully meet the specified constraints, reapplying the constraints to the generator resulted in final outputs that reflected them appropriately.
These results suggest that the generator complements the retrieved layouts, thereby enabling our framework to \textbf{produce layouts aligned with paper structures and faithful to layout constraints}.

\section{Conclusion}
We addressed the problem of understanding and generating poster layouts from scientific papers by constructing SciPostGen, a large-scale dataset of paired papers and poster layouts.
SciPostGen enables systematic analysis and learning of relationships between papers and layouts, as well as reliable evaluation of poster layout generation.
Building on this dataset, we explored a Retrieval-Augmented Poster Generation framework.
Our analyses and experiments showed that paper structures are associated with the number of layout elements in posters, and that the layout retriever can estimate layouts aligned with these structures.
Furthermore, our qualitative analysis suggested that our framework can refine retrieved layouts according to layout constraints typically specified by poster creators.
By introducing SciPostGen, we aimed to bridge data-driven understanding and the generation of poster layouts from scientific papers.
We hope that our work will serve as a foundation for future studies on the generation of scientific posters.

\paragraph{Acknowledgments.}
This work was supported by JST Moonshot R\&D Program, Grant Number JPMJMS2236.

{
    \small
    \bibliographystyle{ieeenat_fullname}
    \bibliography{main}
}
\clearpage
\appendix
\setcounter{page}{1}
\maketitlesupplementary
\section{Dataset Details}
\begin{figure*}[p]
    \centering
    \includegraphics[width=\linewidth]{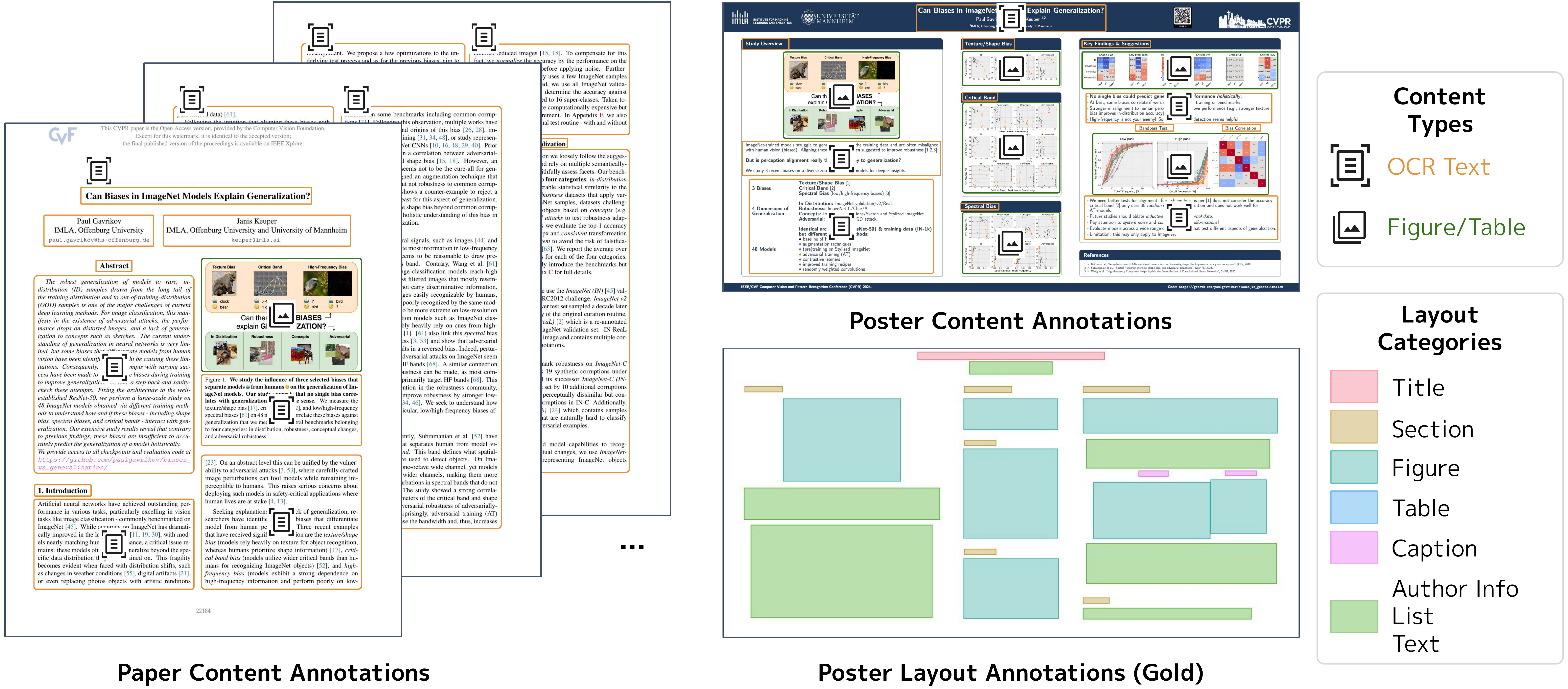}
    \caption{Example of annotations in SciPostGen, including automatically extracted paper and poster annotations and manually corrected poster layout annotations: Paper and poster from \cite{Gavrikov-and-Keuper_2024_CVPR}, licensed under CC BY-SA 4.0.}
    \label{fig:x_dataset_example}
\end{figure*}

\paragraph{Overview.}
Figure~\ref{fig:x_dataset_example} illustrates an example of the annotated components in SciPostGen, a dataset comprising 18,097 pairs of scientific papers and their corresponding posters.

In the main text, we focused on the paper content annotations (e.g., OCR text and figure/table bounding boxes) and the poster layout annotations.
In practice, however, SciPostGen also includes automatically derived poster content annotations.
We omitted these details from the main text to maintain clarity, as they are not directly used in our poster layout generation task.
These poster OCR texts were extracted automatically using Tesseract\footnote{\url{https://github.com/tesseract-ocr/tesseract}}, an open-source OCR engine.

\paragraph{Resources.}
SciPostGen is constructed from open-access papers and their associated posters released by four machine learning and computer vision conferences, as follows:
\begin{tcolorbox}[colback=gray!10, colframe=white, boxrule=0pt, left=2mm, right=2mm, top=1mm, bottom=1mm]
\begin{itemize}
    \item CVPR: \url{https://cvpr.thecvf.com/}
    \item ICLR: \url{https://iclr.cc}
    \item ICML: \url{https://icml.cc}
    \item NeurIPS: \url{https://neurips.cc}
\end{itemize}
\end{tcolorbox}

\paragraph{Comparison with Existing Paper--Poster Datasets.}
Table~\ref{tab:x_dataset_comparisons} summarizes existing datasets that pair scientific papers with their corresponding posters or poster layouts.
While SciPostLayout contains 7,855 poster layouts in total, only 100 of them are paired with their corresponding papers, and thus only these pairs are included in the comparison.
SciPostGen is larger than prior datasets, offering a more substantial foundation for benchmarking poster layout generation and potentially supporting broader poster generation tasks.

\paragraph{Additional Statistics.}
Table~\ref{tab:x_dataset_statistics} reports additional statistics of SciPostGen, which consists of 15,710, 399, and 1,988 pairs in the train, valid, and test splits, respectively.
Word and unique-word statistics are computed from OCR text after lowercasing.

\paragraph{Correlation Analysis on the Train Split.}
To complement the analyses in the main text, we additionally report the correlation results on the SciPostGen train split. This allows us to examine whether the trends observed in the gold layouts (test split) are consistent with those derived from the silver layouts (train split).

\begin{figure*}[p]
    \begin{minipage}[b]{0.51\linewidth}
        \centering
        \includegraphics[keepaspectratio, width=\linewidth]{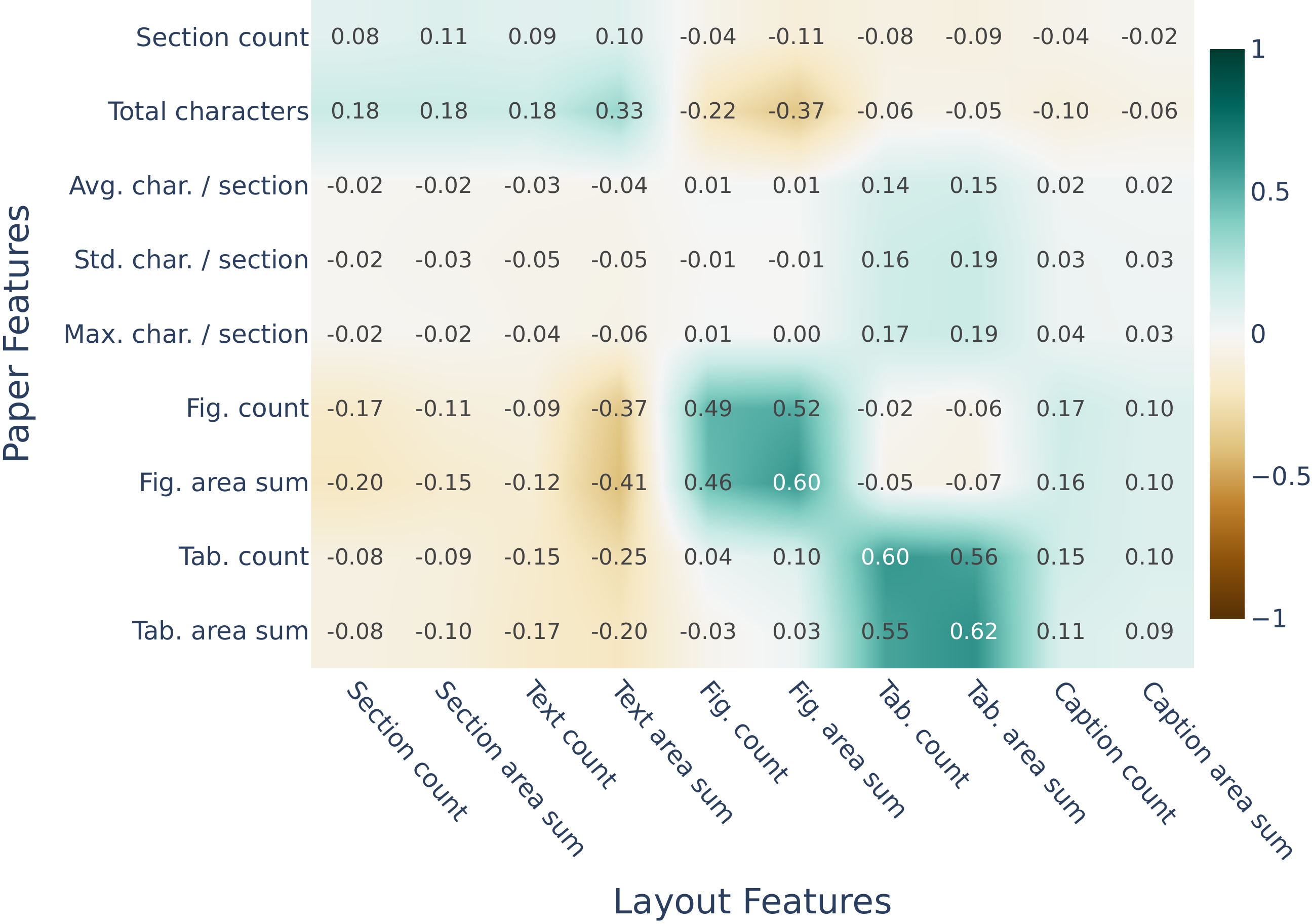}
        \subcaption{Spearman's $\rho$ between paper and layout features}
        \label{fig:x_paper-to-layout}
    \end{minipage}
    \begin{minipage}[b]{0.48\linewidth}
        \centering
        \includegraphics[keepaspectratio, width=\linewidth]{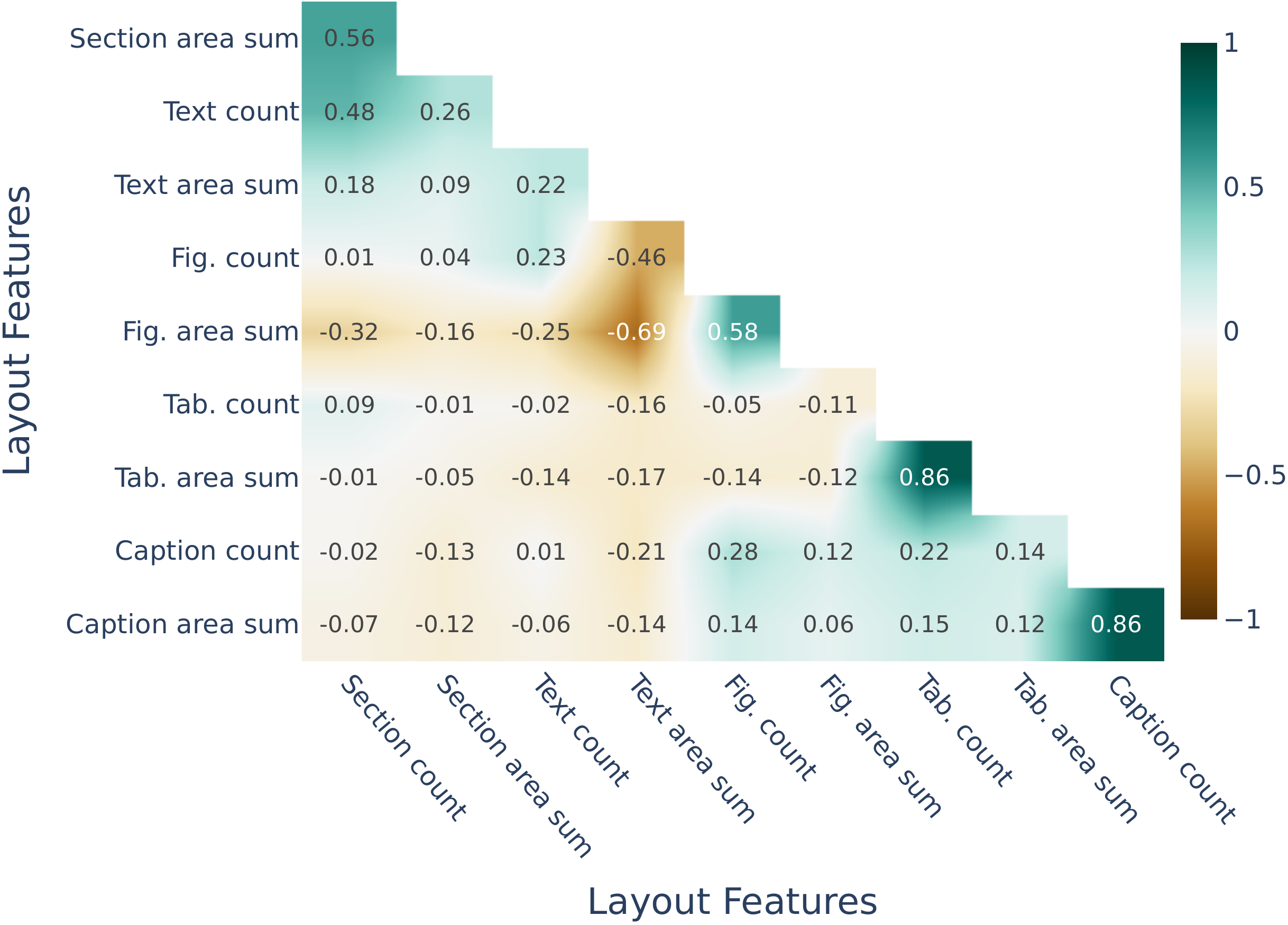}
        \subcaption{Spearman's $\rho$ between layout features}
        \label{fig:x_layout-to-layout}
    \end{minipage}
    \vspace{-2mm}
    \caption{Correlation analyses between paper structures and poster layouts in SciPostGen train split}
    \label{fig:x_corr-matrix}
\end{figure*}

Figure~\ref{fig:x_paper-to-layout} shows the correlation results between paper and layout features on the train split. 
Overall, the trends are consistent with those observed in Section~\ref{sec:3_analysis} based on the gold layouts.

Figure~\ref{fig:x_layout-to-layout} shows the correlation results between layout features on the train split. 
Compared with the gold layout results in Section~\ref{sec:3_analysis}, the silver layouts yield slightly different correlation strengths for certain elements. 
In particular, caption–figure correlations increase ($\rho = 0.28$), and section–text correlations become moderately positive ($\rho = 0.48$). 
These differences may reflect noise in the automatically extracted caption and section annotations.

\begin{table}[t]
    \centering
    \scalebox{1.0}{
    \begin{tabular}{l|c}
            \toprule
            \textbf{Dataset} & \textbf{\#Paper-Poster Pairs} \\
            \cmidrule(lr){0-1}
            NJU‑Fudan~\cite{Qiang-etal_2019} & 85 \\
            Paper2Poster~\cite{Pang-etal_2025_NeurIPS_benchmark} & 100 \\
            P2P eval~\cite{Sun-etal_2025_arXiv} & 121 \\
            SciPostLayout~\cite{Tanaka-etal_2024_BMVC} & 100 \\
            SciPostGen & 18,097 \\
            \bottomrule
    \end{tabular}
    }
    \caption{Comparison of datasets pairing scientific papers with corresponding posters or poster layouts}
    \label{tab:x_dataset_comparisons}
\end{table}

\begin{table}[t]
    \centering
    \begin{minipage}[b]{\linewidth}
        \centering
        \scalebox{1.0}{
        \begin{tabular}{l|ccc}
            \toprule
            \textbf{Paper} & \textbf{Train} & \textbf{Valid}& \textbf{Test} \\
            \midrule
            \# Sections & 109,418 & 2,818 & 14,059 \\
            \# Figures & 81,465 & 2,079 & 10,572 \\
            \# Tables & 56,182 & 1,459 & 7,149 \\
            \cmidrule(lr){0-3}
            Total Chars. (M) & 470.5 & 11.8 & 59.2 \\
            Total Words (M) & 102.4 & 2.5 & 12.7 \\
            Uniq. Words (k) & 314.2 & 39.0 & 98.8 \\
            \bottomrule
        \end{tabular}
        }
        \vspace{0.5mm}
        \subcaption{Paper statistics}
    \end{minipage}

    \vspace{1mm}

    \begin{minipage}[b]{\linewidth}
        \centering
        \scalebox{1.0}{
        \begin{tabular}{l|ccc}
            \toprule
            \textbf{Poster} & \textbf{Train} & \textbf{Valid} & \textbf{Test} \\
            \midrule
            \# Sections & 128,263 & 2,155 & 10,605 \\
            \# Figures & 88,773 & 2,317 & 12,511 \\
            \# Tables & 25,825 & 661 & 3,436 \\
            \cmidrule(lr){0-3}
            Total Chars. (M) & 56.1 & 1.4 & 7.3 \\
            Total Words (M) & 14.0 & 0.3 & 1.8 \\
            Uniq. Words (k) & 278.7 & 27.6 & 78.1 \\
            \bottomrule
        \end{tabular}
        }
        \vspace{0.5mm}
        \subcaption{Poster statistics}
    \end{minipage}
    \caption{Statistics of SciPostGen}
    \label{tab:x_dataset_statistics}
\end{table}


\clearpage
\section{Experimental Details}
\paragraph{Split Usage.}
The train split was used to train the retrieval module. During inference, we queried the trained retriever with papers from the test split and retrieved candidate poster layouts from the train split. 
The valid split was used both for epoch-wise evaluation and hyperparameter selection during the retriever training, and for
ranking layout candidates in the generator.

\paragraph{Data Preprocessing.}
Because the train split does not include the ``Author Info'' and ``List'' categories, we merged them into the ``Text'' category to ensure consistency with the valid and test splits. 
This unified annotation scheme was used both for training the retriever and for evaluating our framework.

\begin{figure}[t]
    \centering
    \includegraphics[keepaspectratio, width=\linewidth]{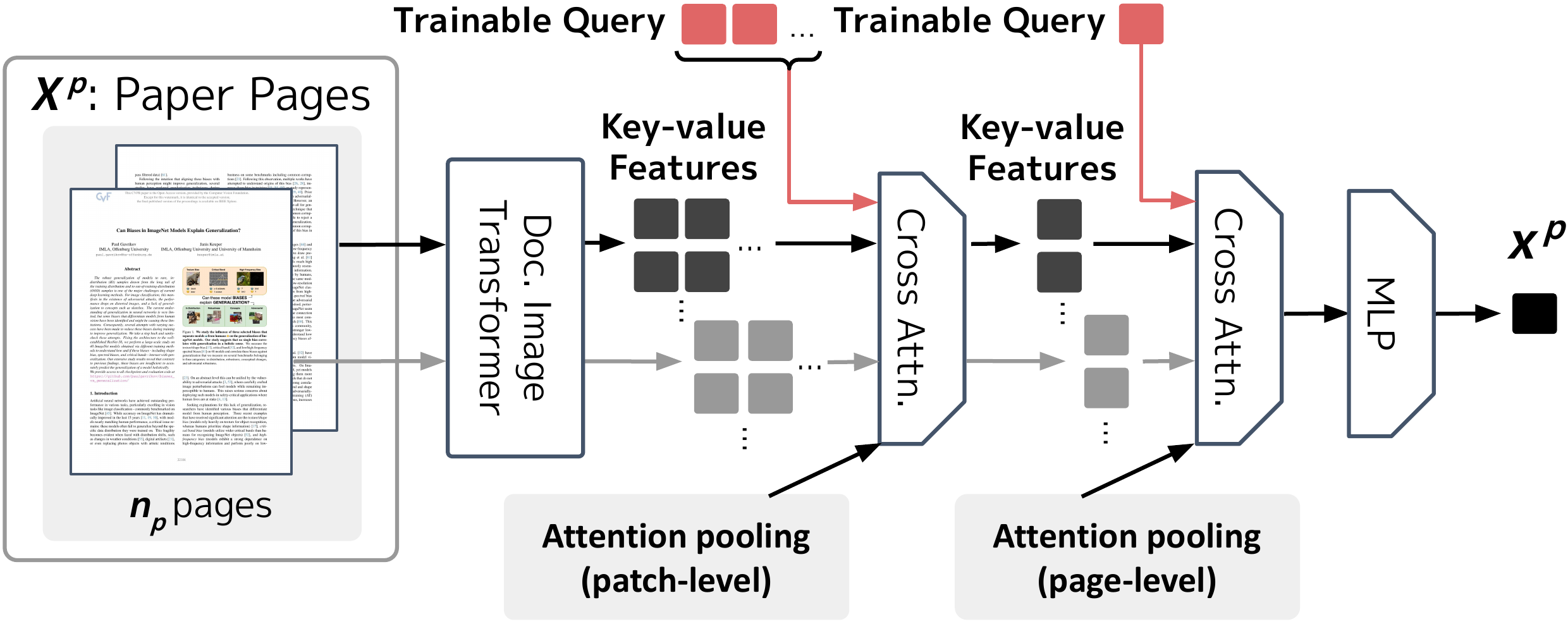}
    \caption{Detailed architecture of the paper encoder}
    \label{fig:x_paper_encoder}
\end{figure}

\paragraph{Implementation Details.}
We use DiT-base\footnote{\texttt{dit-base}: \url{https://huggingface.co/microsoft/dit-base}}
as the backbone encoder for both paper pages and layout images.
Figure~\ref{fig:x_paper_encoder} illustrates the architecture of the paper encoder, which applies patch-level and page-level pooling over the backbone outputs to obtain a paper embedding.
The layout encoder follows a similar structure but does not include page-level pooling.
Table~\ref{tab:x_ret_parameters} summarizes the number of trainable parameters in the
retriever, which contains 180M parameters in total.

The retriever settings are as follows:
\begin{tcolorbox}[floatplacement=H, colback=gray!10, colframe=white, boxrule=0pt, left=2mm, right=2mm, top=1mm, bottom=1mm]
\begin{itemize}
    \item Paper and layout image sizes: $H = W = 224$
    \item Number of paper pages: $n_p = 8$
    \item Paper and layout embedding dimension: $d = 256$
\end{itemize}
\end{tcolorbox}

We trained the retriever on 4$\times$NVIDIA A100 GPUs in 12 hours by AdamW~\cite{Loshchilov-and-Hutter_2018_ICLR}. 
The training hyperparameters are as follows:
\begin{tcolorbox}[floatplacement=H, colback=gray!10, colframe=white, boxrule=0pt, left=2mm, right=2mm, top=1mm, bottom=1mm]
\begin{itemize}
    \item Batch size: $N = 128$
    \item Temperature parameter: $\tau = 0.07$
    \item Number of epochs: 20
    \item Learning rate: $\{1\times10^{-4}, 1\times10^{-5}, 1\times10^{-6}\}$
    \item Weight decay: 0.01
    \item Scheduler: Linear Warmup Cosine Annealing
    \item Warmup ratio: 0.1
\end{itemize}
\end{tcolorbox}

We used two large language models for layout generation:
GPT-5-mini\footnote{\texttt{gpt-5-mini-2025-08-07}: \url{https://platform.openai.com/docs/models/gpt-5-mini}}
and GPT-5\footnote{\texttt{gpt-5-2025-08-07}: \url{https://platform.openai.com/docs/models/gpt-5}}.
For cost considerations, we set the reasoning effort to low for GPT-5-mini and minimal for GPT-5.
Under these settings, the API cost for generating layouts for the 1,988 samples in the
SciPostGen test split was \$8--10 with GPT-5-mini and \$35--40 with GPT-5.

Table~\ref{tab:prompt_uncond} and Table~\ref{tab:prompt_cond} show the example prompts used in the automatic and semi-automatic poster generation settings, respectively.
In both settings, the generator receives the following paper structures:
\begin{tcolorbox}[floatplacement=H, colback=gray!10, colframe=white, boxrule=0pt, left=2mm, right=2mm, top=1mm, bottom=1mm]
\begin{itemize}
    \item Number of sections, captions, figures, and tables
    \item Characters of title, authors, abstract, and each section
    \item Aspect ratios of figures and tables
\end{itemize}
\end{tcolorbox}

\begin{table}[t]
    \centering
    \scalebox{0.73}{
        \begin{tabular}{l|ccc}
        \toprule
        \textbf{Component} & \textbf{Backbone} & \textbf{Other Params} & \textbf{Total} \\
        \cmidrule(lr){0-3}
        Paper Encoder & 85M & 6M & 91M \\
        Layout Encoder & 85M & 4M & 89M \\
        \bottomrule
        \end{tabular}
    }
    \caption{Number of trainable parameters in each encoder of the retriever: The retriever contains 180M trainable parameters in total}
    \label{tab:x_ret_parameters}
\end{table}

\begin{table*}[p]
\centering
\begin{tcolorbox}[fontupper=\scriptsize, title=\small Prompt for Automatic Generation with Paper Structure]
\begin{verbatim}
Task: Layout generation conditioned on paper statistics. 
Please use layout element types: Title, Section, Text, Figure, Table, Caption.

Here is example 1.
Paper Info: 
    #Paper element counts
        Section: 9, Caption: 6, Figure: 7
    #Paper element lengths in characters
        Title length: 69, Author length: 625, Abstract length: 961, 1 Introduction length: 7242 ...
    #Paper figure/table aspect ratios (width/height)
        Figure 1: 1.92, Figure 2: 4.86 ...
Input: 
    <html><body>
        <div class='canvas' style='left: 0px; top: 0px; width: 5120px; height: 2560px'></div>
    </body></html>
Output: 
    <html><body>
        <div class='canvas' style='left: 0px; top: 0px; width: 3456px; height: 2304px'></div>
        <div class='Title' style='left: 57px; top: 61px; width: 3306px; height: 76px;'></div>
        ...
    </html></body>

Here is example 2.
...

Please generate a layout.
Paper Info: ...
Input: ...
Output: 
\end{verbatim}
\end{tcolorbox}
\caption{Example prompt used in the automatic poster generation setting}
\label{tab:prompt_uncond}
\end{table*}

\begin{table*}[p]
\centering
\begin{tcolorbox}[fontupper=\scriptsize, title=\small Prompt for Semi-Automatic Generation with Paper Structure and Layout Constraints]
\begin{verbatim}
Task: Layout completion conditioned on partial layout and paper statistics.
Please use layout element types: Title, Section, Text, Figure, Table, Caption.

Here is example 1.
Paper Info: 
    #Paper element counts
        Section: 9, Caption: 6, Figure: 7
    #Paper element lengths in characters
        Title length: 69, Author length: 625, Abstract length: 961, 1 Introduction length: 7242 ...
    #Paper figure/table aspect ratios (width/height)
        Figure 1: 1.92, Figure 2: 4.86 ...
Input: 
    <html><body>
        <div class='canvas' style='left: 0px; top: 0px; width: 5120px; height: 2560px'></div>
        <div class='Figure' style='left: 1522px; top: 1306px; width: 1761px; height: 778px;'></div>
        <div class='Text' style='left: 57px; top: 1944px; width: 2298px; height: 544px;'></div>
    </body></html>
Output: 
    <html><body>
        <div class='canvas' style='left: 0px; top: 0px; width: 5120px; height: 2560px'></div>
        <div class='Title' style='left: 1265px; top: 26px; width: 2561px; height: 299px;'>
        ...
        <div class='Figure' style='left: 1522px; top: 1306px; width: 1761px; height: 778px;'></div>
        <div class='Text' style='left: 57px; top: 1944px; width: 2298px; height: 544px;'></div>
        ...
    </html></body>

Here is example 2.
...

Please generate a layout.
Paper Info: ...
Input: ...
Output: 
\end{verbatim}
\end{tcolorbox}
\caption{Example prompt used in the semi-automatic poster generation setting}
\label{tab:prompt_cond}
\end{table*}

\clearpage
\section{Additional Results}
\paragraph{Comparison under the Oracle Setting.}
We compare the predicted layouts with the oracle setting, which consists of the gold layouts and retriever upper bounds computed for each test sample by selecting the best-matching silver layout from the training pool in terms of mIoU or TC$_{std}$.
For reference-free evaluation, we report Overlap and Alignment. Overlap measures the extent to which layout elements undesirably intersect, while Alignment measures how well their edges and centers are aligned.

Figure~\ref{tab:x_comparison_upper-and-gold} shows that the layouts generated by GPT-5 show higher Alignment scores while achieving lower Overlap than the retrieved layouts. 
The gold layouts, however, do not show high Alignment (0.065), suggesting that strong layout alignment is not necessarily a characteristic of human-designed scientific posters.

The retriever upper bound corresponds to 0.347 in mIoU and 0.390 in TC$_{std}$, computed by selecting the best-matching training layout for each test sample.
Although our predicted layouts do not reach this upper bound, these oracle values show the potential of Retrieval-Augmented Poster Layout generation. 

\paragraph{Ablation Study of the Retriever.}
Figure~\ref{tab:x_ablation_for_poster_inputs} shows the effect of replacing layout images with full poster canvases as the input to the layout encoder. 
We observe that using posters results in consistently lower performance across all metrics compared with using the layout
images. 
This suggests that visual elements such as colors and decorative styling do not help the retriever estimate layouts that align with the corresponding papers.

\begin{figure}[b]
    \begin{minipage}[b]{0.48\linewidth}
        \centering
        \includegraphics[keepaspectratio, height=0.137\textheight]{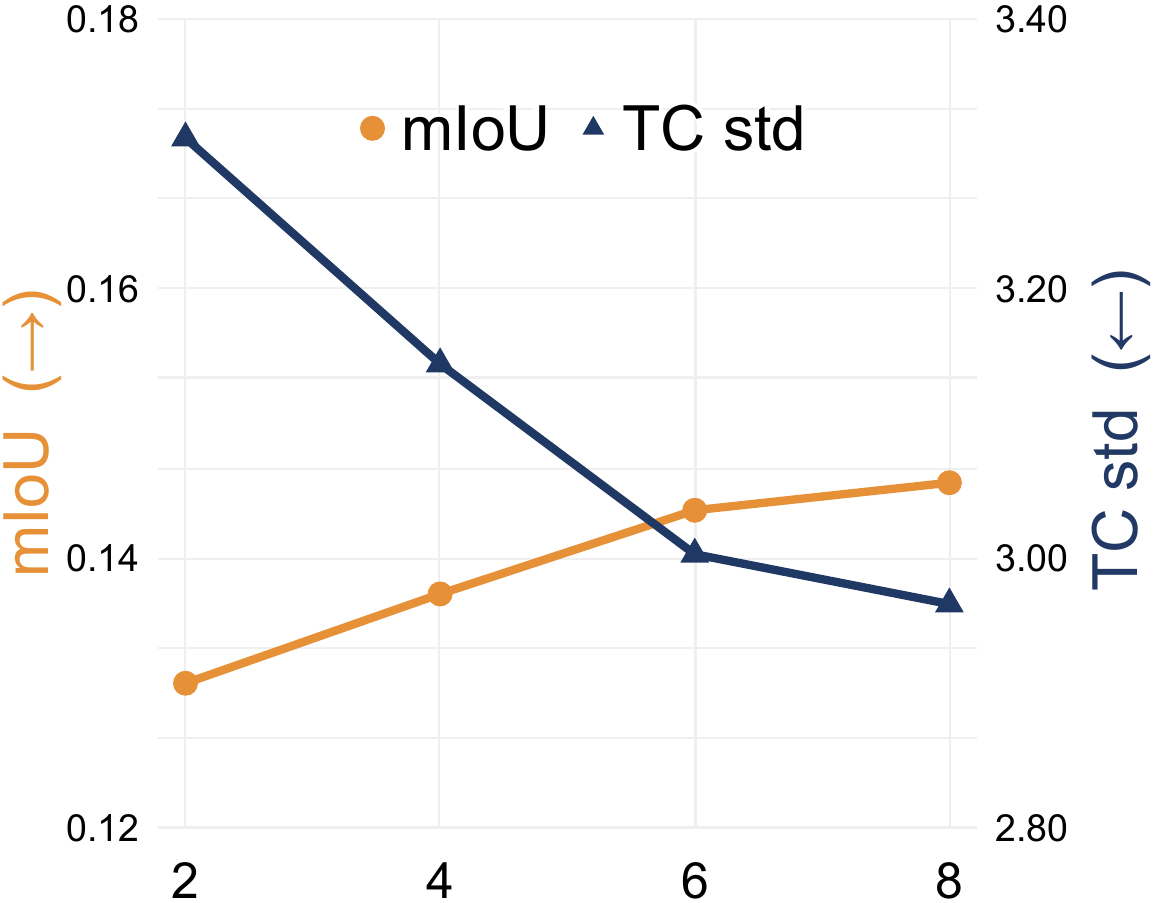}
        \subcaption{Paper pages}
        \label{fig:x_ablation_pages}
    \end{minipage}
    \begin{minipage}[b]{0.51\linewidth}
        \centering
        \includegraphics[keepaspectratio, height=0.137\textheight]{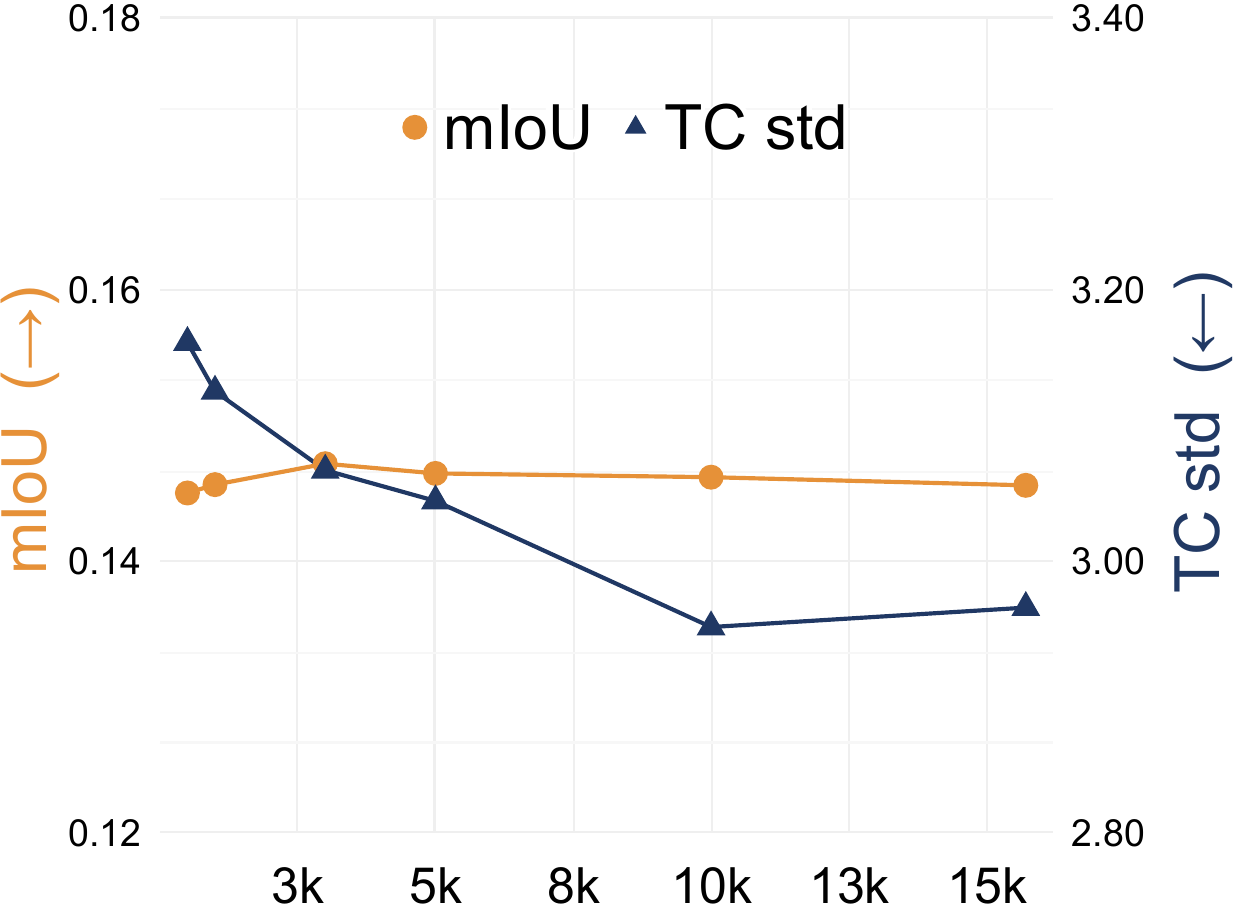}
        \subcaption{Retrieval pool size}
        \label{fig:x_ablation_retrieval-pool}
    \end{minipage}
    \caption{Ablation results for the retriever: (a) Effect of changing the number of input paper pages $n_p$ by training the retriever. (b) Effect of varying the retrieval pool size during inference ($n_p=8$)}
\end{figure}

We reduce the number of input pages $n_p$ from the default 8 to 6, 4, and 2, training and evaluating the retriever for each setting. 
Figure~\ref{fig:x_ablation_pages} shows that retrieval performance consistently improves as $n_p$ increases: mIoU increases while TC$_{std}$ decreases. 
These results indicate that incorporating more pages is beneficial for retrieval and suggest that the attention pooling components effectively aggregate global information across the entire paper.

Figure~\ref{fig:x_ablation_retrieval-pool} shows how retrieval performance changes as the pool size is reduced, based on the retriever trained with $n_p = 8$.
The mIoU remains stable across different pool sizes, whereas TC$_{std}$ improves as the pool size increases but saturates around a pool size of 10k.

\begin{figure}[b]
    \centering
    \includegraphics[keepaspectratio, width=\linewidth]{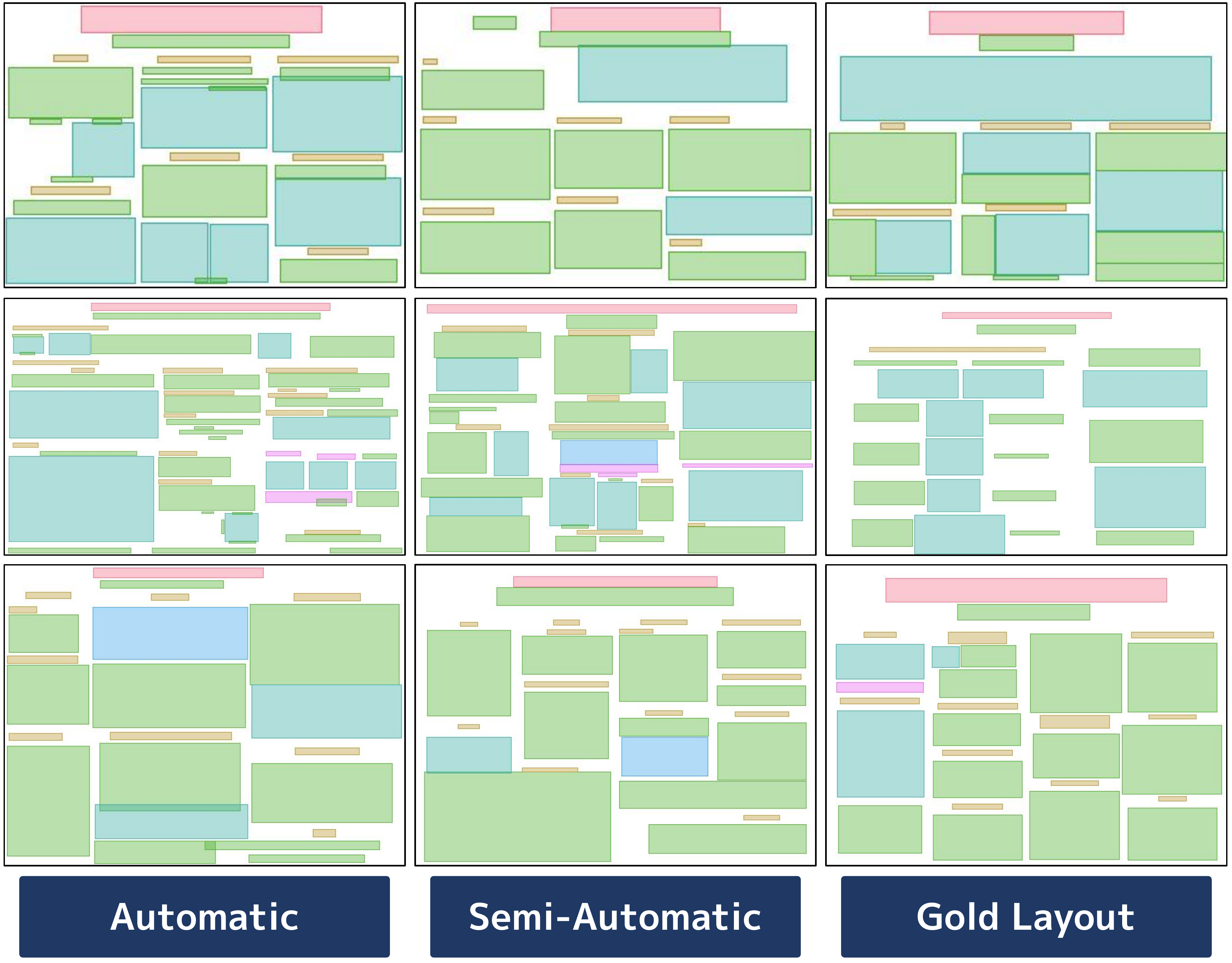}
    \caption{Failure cases of the retriever: we show the retrieved top-1 layouts under the automatic and semi-automatic settings}
    \label{fig:x_failure_cases}
\end{figure}

\paragraph{Failure Cases of the Retriever.}
Figure~\ref{fig:x_failure_cases} shows failure cases of the retriever. 
Such cases often involve layouts that span multiple columns or adopt unique element arrangements. 
Handling these layouts requires combining retrieval with the generator rather than relying on the retriever alone.

\begin{table}[b]
\centering
    \scalebox{0.73}{
        \begin{tabular}{l|cccc}
        \toprule
        \textbf{Approach} & \textbf{Overall} & \textbf{Alignment}$^{\dagger}$ &\textbf{mIoU $\uparrow$} & \textbf{TC}$_{std}$ $\downarrow$ \\
        \cmidrule(lr){0-4}
        \multicolumn{5}{c}{\textbf{Unconditional Setting} w/ \textbf{GPT-5}} \\
        Retrieved Top-3 (Avg.) & 0.027 & 0.077 & 0.145 & 2.965 \\
        $\hookrightarrow$ Paper Structures & 0.065 & 0.001 & 0.159 & 3.128 \\
        \cmidrule(lr){0-4}
        \multicolumn{5}{c}{\textbf{Conditional Setting} w/ \textbf{GPT-5}} \\
        Retrieved Top-3 (Avg.) & 0.026 & 0.075 & 0.196 & 2.873 \\
        $\hookrightarrow$ Paper Structures & 0.037 & 0.011 & 0.238 & 2.941 \\
        \cmidrule(lr){0-4}
        Retriever Upper Bound & --- & --- & 0.347 & 0.390 \\
        Gold Layout & 0.003 & 0.065 & --- & --- \\
        \bottomrule
        \end{tabular}
    }
    \caption{Comparison of retrieved layouts, retriever upper bound, and gold layouts: ``Overall'' and ``Alignment'' metrics are better when their values are closer to those of the gold layouts. Values in the columns marked with $\dagger$ are scaled by a factor of 100.}
    \label{tab:x_comparison_upper-and-gold}
\end{table}

\begin{table}[b]
\centering
    \scalebox{0.73}{
        \begin{tabular}{l|cccc}
        \toprule
        \textbf{Poster Input} & \makebox[1.4cm]{\textbf{mIoU} $\uparrow$} & \makebox[1.4cm]{\textbf{LTSim} $\uparrow$} & \textbf{TC}$_{mean}$ $\rightarrow$ & \textbf{TC}$_{std}$ $\downarrow$ \\
        \cmidrule(lr){0-4}
        Layout & 0.145 & 0.651 & -0.057 & 2.965 \\
        Poster & 0.137 & 0.646 & -0.705 & 3.456 \\
        \bottomrule
        \end{tabular}
    }
    \caption{Results of retriever training with different poster inputs}
    \label{tab:x_ablation_for_poster_inputs}
\end{table}


\clearpage

\paragraph{Comparison Across LLM Generators.}
Tables~\ref{tab:x_comparison_under_unconditional} and \ref{tab:x_comparison_under_conditional} compare different open-source LLM generators: GPT-OSS-120b\footnote{\url{https://huggingface.co/openai/gpt-oss-120b}}, Qwen3-32B\footnote{\url{https://huggingface.co/Qwen/Qwen3-32B}}, and Qwen3-Coder-30B\footnote{\url{https://huggingface.co/Qwen/Qwen3-Coder-30B-A3B-Instruct}}, in the automatic and semi-automatic settings~\cite{openai_2025_GPT-OSS, Qwen-Team_2025_Qwen3}. 
We observe that performance varies across models in the automatic setting, indicating that generation quality depends on the underlying LLM. 
In the semi-automatic setting, the best-performing open-source LLMs (Qwen3-32B and Qwen3-Coder-30B) achieve mIoU values close to GPT-5-mini. 
This suggests that the evaluated open-source LLMs can effectively follow user-specified layout constraints.

\paragraph{Ablation Study on Paper Structures.}
Table~\ref{tab:x_ablation_for_paper_inputs} shows the effect of different paper structures in the automatic setting, where GPT-5 is used as the generator.
We observe that using all structures together yields higher mIoU than using any single structure, suggesting that these structures provide complementary information.

\section{Limitations}
We acknowledge that SciPostGen is limited to computer science papers and primarily landscape-format posters, which may restrict the diversity of represented domains and layout styles.
Future extensions to encompass broader research domains and poster orientations could mitigate this limitation.
Our framework focuses on layout generation and does not address multimodal summarization.
We also do not include comparisons with existing layout generation models, since generating layouts from lengthy and structured scientific papers remains an exploratory problem, and existing models are not directly applicable without substantial adaptation.

\begin{table}[t]

    \begin{minipage}[t]{\linewidth}
        \centering
        \scalebox{0.74}{
            \begin{tabular}{l|cccc}
                \toprule
                \textbf{Approach} & \textbf{mIoU} $\uparrow$ & \textbf{LTSim} $\uparrow$ & \textbf{TC}$_{mean}$ $\rightarrow$ & \textbf{TC}$_{std}$ $\downarrow$ \\
                \cmidrule(lr){0-4}
                Retrieved Top-3 (Avg.) & 0.145 & 0.651 & -0.057 & 2.965 \\
                $\hookrightarrow$ w/o Paper Structures & 0.133 & 0.638 & 1.484 & 2.816 \\
                $\hookrightarrow$ w/\phantom{o} Paper Structures & 0.139 & 0.630 & 0.444 & 2.905 \\
                \bottomrule
            \end{tabular}
        }
        \vspace{0.5mm}
        \label{tab:x_comp_uncond_qwen3-coder-32b}
        \subcaption{\textbf{Qwen3-32b}}
    \end{minipage}

    \vspace{1mm}

    \begin{minipage}[t]{\linewidth}
        \centering
        \scalebox{0.74}{
            \begin{tabular}{l|cccc}
                \toprule
                \textbf{Approach} & \textbf{mIoU} $\uparrow$ & \textbf{LTSim} $\uparrow$ & \textbf{TC}$_{mean}$ $\rightarrow$ & \textbf{TC}$_{std}$ $\downarrow$ \\
                \cmidrule(lr){0-4}
                Retrieved Top-3 (Avg.) & 0.145 & 0.651 & -0.057 & 2.965 \\
                $\hookrightarrow$ w/o Paper Structures & 0.138 & 0.646 & 0.946 & 2.715 \\
                $\hookrightarrow$ w/\phantom{o} Paper Structures & 0.131 & 0.645 & 0.095 & 2.826 \\
                \bottomrule
            \end{tabular}
        }
        \vspace{0.5mm}
        \subcaption{\textbf{Qwen3-Coder-30b}}
        \label{tab:x_comp_uncond_qwen-coder-30b}
    \end{minipage}
    
    \vspace{1mm}

    \begin{minipage}[t]{\linewidth}
        \centering
        \scalebox{0.74}{
            \begin{tabular}{l|cccc}
                \toprule
                \textbf{Approach} &\textbf{mIoU} $\uparrow$ & \textbf{LTSim} $\uparrow$ & \textbf{TC}$_{mean}$ $\rightarrow$ & \textbf{TC}$_{std}$ $\downarrow$ \\
                \cmidrule(lr){0-4}
                Retrieved Top-3 (Avg.) & 0.145 & 0.651 & -0.057 & 2.965 \\
                $\hookrightarrow$ w/o Paper Structures & 0.120 & 0.617 & 1.596 & 3.272 \\
                $\hookrightarrow$ w/\phantom{o} Paper Structures & 0.127 & 0.612 & -0.175 & 3.311 \\
                \bottomrule
            \end{tabular}
        }
        \vspace{0.5mm}
        \subcaption{\textbf{GPT-OSS-120b}}
        \label{tab:x_comp_uncond_gpt-oss-120b}
    \end{minipage}
    \vspace{-3mm}
    \caption{Performance comparison across different LLM generators in automatic generation setting}
    \label{tab:x_comparison_under_unconditional}
\end{table}

\begin{table}[t]    
    \begin{minipage}[t]{\linewidth}
        \centering
        \scalebox{0.74}{
            \begin{tabular}{l|cccc}
                \toprule
                \textbf{Approach} & \textbf{mIoU} $\uparrow$ & \textbf{LTSim} $\uparrow$ & \textbf{TC}$_{mean}$ $\rightarrow$ & \textbf{TC}$_{std}$ $\downarrow$ \\
                \cmidrule(lr){0-4}
                Retrieved Top-3 (Avg.) & 0.196 & 0.662 & 0.194 & 2.873 \\
                $\hookrightarrow$ w/o Paper Structures & 0.214 & 0.654 & 1.185 & 2.598 \\
                $\hookrightarrow$ w/\phantom{o} Paper Structures & 0.206 & 0.650 & 0.362 & 2.823\\
                \bottomrule
            \end{tabular}
        }
        \vspace{0.5mm}
        \label{tab:x_comp_cond_qwen3-coder-32b}
        \subcaption{\textbf{Qwen3-32b}}
    \end{minipage}

    \vspace{1mm}

    \begin{minipage}[t]{\linewidth}
        \centering
        \scalebox{0.74}{
            \begin{tabular}{l|cccc}
                \toprule
                \textbf{Approach} & \textbf{mIoU} $\uparrow$ & \textbf{LTSim} $\uparrow$ & \textbf{TC}$_{mean}$ $\rightarrow$ & \textbf{TC}$_{std}$ $\downarrow$ \\
                \cmidrule(lr){0-4}
                Retrieved Top-3 (Avg.) & 0.196 & 0.662 & 0.194 & 2.873 \\
                $\hookrightarrow$ w/o Paper Structures & 0.214 & 0.660 & 0.979 & 2.650 \\
                $\hookrightarrow$ w/\phantom{o} Paper Structures & 0.197 & 0.658 & 0.178 & 2.732 \\
                \bottomrule
            \end{tabular}
        }
        \vspace{0.5mm}
        \label{tab:x_comp_cond_qwen3-coder-30b}
        \subcaption{\textbf{Qwen3-Coder-30b}}
    \end{minipage}
    
    \vspace{1mm}

    \begin{minipage}[t]{\linewidth}
        \centering
        \scalebox{0.74}{
            \begin{tabular}{l|cccc}
                \toprule
                \textbf{Approach} & \textbf{mIoU} $\uparrow$ & \textbf{LTSim} $\uparrow$ & \textbf{TC}$_{mean}$ $\rightarrow$ & \textbf{TC}$_{std}$ $\downarrow$ \\
                \cmidrule(lr){0-4}
                Retrieved Top-3 (Avg.) & 0.196 & 0.662 & 0.194 & 2.873 \\
                $\hookrightarrow$ w/o Paper Structures & 0.199 & 0.631 & 2.119 & 3.050 \\
                $\hookrightarrow$ w/\phantom{o} Paper Structures & 0.196 & 0.610 & -0.674 & 3.816 \\
                \bottomrule
            \end{tabular}
        }
        \vspace{0.5mm}
        \label{tab:x_comp_cond_gpt-oss-120b}
        \subcaption{\textbf{GPT-OSS-120b}}
    \end{minipage}
    \vspace{-3mm}
    \caption{Performance comparison across different LLM generators in semi-automatic generation setting}
    \label{tab:x_comparison_under_conditional}
\end{table}
\begin{table}[t]
\centering
    \scalebox{0.74}{
        \begin{tabular}{l|cccc}
        \toprule
        \textbf{Paper Input} & \makebox[1.4cm]{\textbf{mIoU} $\uparrow$} & \makebox[1.4cm]{\textbf{LTSim} $\uparrow$} & \textbf{TC}$_{mean}$ $\rightarrow$ & \textbf{TC}$_{std}$ $\downarrow$ \\
        \cmidrule(lr){0-4}
        Retrieved Top-3 (Avg.) & 0.145 & 0.651 & -0.057 & 2.965 \\
        $\hookrightarrow$  Paper Structure & 0.159 & 0.642 & 0.228 & 3.128 \\
        $\hookrightarrow$  Element Counts & 0.152 & 0.643 & -0.082 & 3.298 \\
        $\hookrightarrow$  Characters & 0.156 & 0.656 & 0.785 & 2.694 \\
        $\hookrightarrow$  Aspect Ratios & 0.148 & 0.631 & 0.638 & 3.527 \\
        \bottomrule
        \end{tabular}
    }
    \caption{Results of framework with different paper inputs in automatic setting using GPT-5 as the generator.}
    \label{tab:x_ablation_for_paper_inputs}
\end{table}

\end{document}